%% file: main.tex
\documentclass{article}
\usepackage{microtype}
\usepackage{graphicx}
\usepackage{subfigure}
\usepackage{booktabs} 

\usepackage{hyperref}
\usepackage{enumitem}

\usepackage[ruled,vlined]{algorithm2e}

\usepackage[accepted]{icml2024}

\usepackage{amsmath}
\usepackage{amssymb}
\usepackage{mathtools}
\usepackage{amsthm}
\usepackage{multirow}
\usepackage{hyperref}
\usepackage{bm} 
\usepackage[capitalize,noabbrev]{cleveref}

\theoremstyle{plain}

\theoremstyle{definition}

\theoremstyle{remark}

\usepackage[textsize=tiny]{todonotes}
\usepackage{listings}
\icmltitlerunning{PAPM: A Physics-aware Proxy Model for Process Systems}

\begin{document}

\twocolumn[
\icmltitle{PAPM: A Physics-aware Proxy Model for Process Systems}



\begin{icmlauthorlist}
\icmlauthor{Pengwei Liu}{xxx}
\icmlauthor{Zhongkai Hao}{yyy}
\icmlauthor{Xingyu Ren}{xxx}
\icmlauthor{Hangjie Yuan}{xxx}
\icmlauthor{Jiayang Ren}{zzz}
\icmlauthor{Dong Ni}{xxx}
\end{icmlauthorlist}
\icmlaffiliation{xxx}{Zhejiang University, Hangzhou, Zhejiang, China}
\icmlaffiliation{yyy}{Tsinghua University, Beijing, China}
\icmlaffiliation{zzz}{University of British Columbia, Vancouver, BC, Canada}
\icmlcorrespondingauthor{Dong Ni}{dni@zju.edu.cn}
\icmlkeywords{Process systems modeling, Physics-informed deep learning, Temporal-spatial stepping method, Out-of-sample generalization}

\vskip 0.3in
]



\printAffiliationsAndNotice{}  

\begin{abstract}
In the context of proxy modeling for process systems, traditional data-driven deep learning approaches frequently encounter significant challenges, such as substantial training costs induced by large amounts of data, and limited generalization capabilities. 
As a promising alternative, physics-aware models incorporate partial physics knowledge to ameliorate these challenges.
Although demonstrating efficacy, they fall short in terms of exploration depth and universality.
To address these shortcomings, we introduce a \textbf{p}hysics-\textbf{a}ware \textbf{p}roxy \textbf{m}odel (\textbf{PAPM}) that fully incorporates partial prior physics of process systems, which includes multiple input conditions and the general form of conservation relations, resulting in better out-of-sample generalization.
Additionally, PAPM contains a holistic temporal-spatial stepping module for flexible adaptation across various process systems.
Through systematic comparisons with state-of-the-art pure data-driven and physics-aware models across five two-dimensional benchmarks in nine generalization tasks, PAPM notably achieves an average performance improvement of 6.7\%, while requiring fewer FLOPs, and just 1\% of the parameters compared to the prior leading method. 
Code is available at \href{https://github.com/pengwei07/PAPM}{https://github.com/pengwei07/PAPM}.
\end{abstract}

\section{Introduction}
\begin{figure*}[t]
\centering
\includegraphics[width=0.9\textwidth]{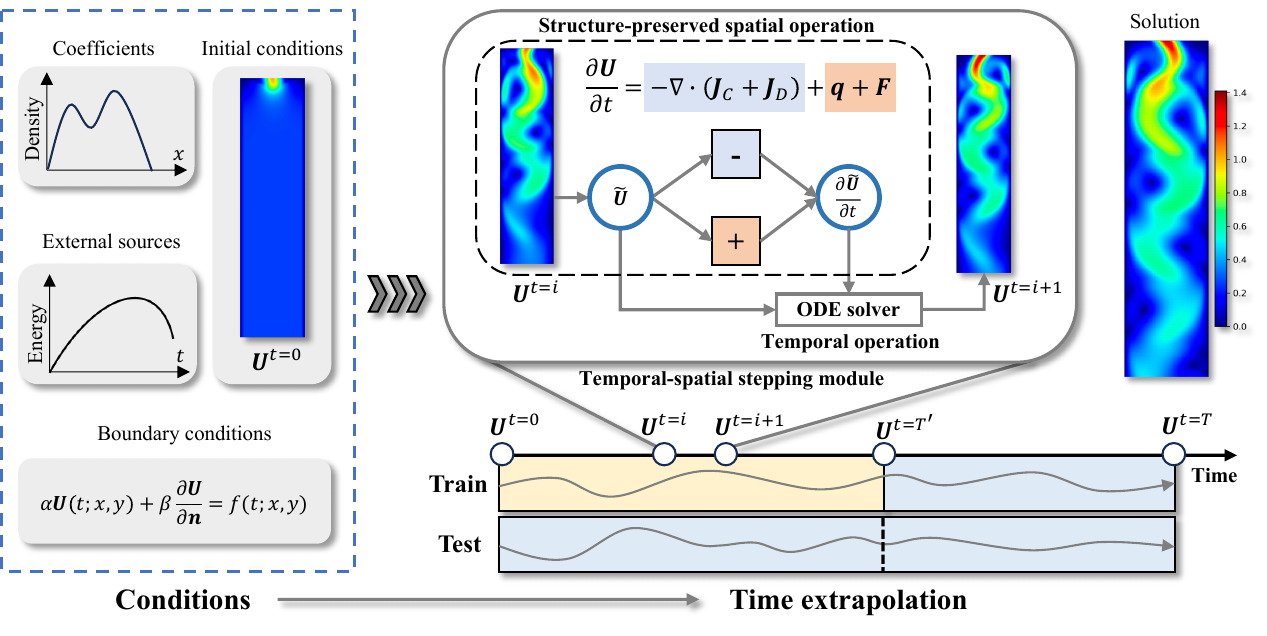}
\vspace{-4mm}
\caption{Overview of the PAPM's pipeline. The model takes the multiple conditions of process systems for time extrapolation and outputs solutions at an arbitrary time point. The core is the temporal-spatial stepping module (TSSM) $(\boldsymbol{U}^{t=i}\rightarrow\boldsymbol{U}^{t=i+1})$. Spatially, a structure-preserved operation aligns with the specific equation characteristics of different process systems. Temporally, it utilizes a continuous-time modeling framework through an ODE solver.
}\label{pipline}
\vspace{-2mm}
\end{figure*}

From molecular dynamics to turbulent flows, process systems are essential in numerous scientific and engineering domains~\citep{cameron2001process}. Computational modeling and simulation are crucial for understanding their complex temporal-spatial dynamics, enabling precise predictions and informed decisions across various fields.
However, these valuable insights are provided by traditional numerical simulations, which are often computationally intensive, especially in scenarios necessitating frequent model queries like reverse engineering forward simulation~\citep{dijkstra2021predictive} and optimization design~\citep{gramacy2020surrogates}.
Recent advancements in data-driven methods have paved the way to tackle computational challenges more effectively~\citep{lu2019deeponet, li2020fourier, kochkov2021machine, stachenfeld2021learned, hao2023gnot}. 
As shown in Fig.~\ref{pipline} (left), these methods input multiple conditions of process systems to output time-dependent solutions, serving as the proxy model for process systems. 
Through adopting a supervised learning-from-data paradigm, remarkable advancements in calculation speeds have been achieved—several orders of magnitude faster than traditional numerical simulations. This has led to significant savings in computational costs.
While data-driven methods are powerful, they face two primary limitations: 
\textbf{1)} a dependence on extensive labeled datasets, which contrasts sharply with the high computational costs of numerical simulations, 
and \textbf{2)} a presumption of train-test uniformity that leads to poor generalization, especially in out-of-sample scenarios like time extrapolation. 
This poor generalization arises from an overemphasis on the inductive biases of network architectures based on labels, rather than a strict adherence to fundamental physical laws~\citep{li2021physicsinformed, brandstetter2022message}.

To ameliorate high training costs and limited generalizability, a more promising strategy, known as physics-informed deep learning (PIDL), involves prior knowledge, such as fundamental physical laws, into neural networks (NNs). This integration enhances the sample efficiency and generalizability of NNs, proving particularly vital in scenarios with limited labeled data~\citep{li2021physicsinformed, hao2022physics, meng2022physics, cuomo2022scientific}.
One of the typical methods considers complete physical laws as the loss function of NNs to construct proxy models, such as PINNs~\citep{raissi2019physics} for specific conditions, PI-DeepONet~\citep{wang2021learning}, and PINO~\citep{li2021physicsinformed} for multiple sets of conditions. 
However, the real-world applicability of these methods is limited by an incomplete understanding of the underlying physics of specific process systems, making it challenging to derive complete physics laws. 

Another common approach, termed as ``\textbf{physics-aware}" models, offers a promising solution for this scenario by incorporating partial prior knowledge alongside a small amount of labeled data.  Considering the relationship between equations and their numerical schemes, the physics-aware methods convert partial prior physics laws into the corresponding numerical schemes, and embeddings them into the network structure~\citep{long2018pde, seo2020physics, huang2023neuralstagger, akhare2023physics, rao2023encoding, huang2023learning, dmitrii2023Neural, pestourie2023physics, liu2024multi}. 
With these partial prior physics laws, physics-aware models only need a small number of labels to obtain excellent out-of-sample generalization performance. 
However, these methods focus primarily on spatial derivatives, and often neglect integral aspects such as conservation laws and constitutive relations. 
Consequently, they do not fully leverage prior physics knowledge, leading to unreliable solutions.
Besides, these methods are generally tailored for a specific process system with limited universality. 

Recognizing that process system modeling often requires the incorporation of conservation relations grounded in diffusion, convection, and source flows, this work aims to integrate this general physics law into the network architecture. By reinforcing the inductive biases in this manner, we can achieve better out-of-sample generalization. 
Additionally, as different process systems correspond to specific conservation or constitutive equations based on inherent system characteristics~\citep{cameron2001process, takamoto2022pdebench, hao2023pinnacle}, it is beneficial to identify both similarities and differences among process systems. Such an approach can offer a general temporal-spatial stepping module to combine various process systems flexibly.

As illustrated in Fig.~\ref{pipline}, we propose a \textbf{p}hysics-\textbf{a}ware \textbf{p}roxy \textbf{m}odel (\textbf{PAPM}) for process systems, which incorporates multiple conditions to output solutions at arbitrary time points.
PAPM fully utilizes partial prior knowledge, including multiple conditions, and the general form of conservation relations, alongside a small amount of label data, to model the dynamics of systems through the proposed temporal-spatial stepping module $(\boldsymbol{U}^{t=i} \rightarrow \boldsymbol{U}^{t=i+1})$.
Notably, PAPM leverages the direction of data flow based on this general form, a distinction often overlooked by other physics-aware methods.
Furthermore, PAPM focuses on out-of-sample scenarios, such as time extrapolation, aligning with the capabilities of alternative methods.

The core contributions of this work are: 
\begin{itemize}[itemsep=1pt,topsep=0pt,leftmargin=*]
\item The proposal of PAPM, a versatile physics-aware architecture design that fully incorporates partial prior knowledge such as multiple conditions, and the general form of conservation relations. This design proves to be superior in both training efficiency and out-of-sample generalizability.
\item The introduction of a holistic temporal-spatial stepping module (TSSM) for flexible adaptation across various process systems. It aligns with the distinct equation characteristics of different process systems by employing stepping schemes via temporal and spatial operations, whether in physical or spectral space.
\item A systematic evaluation of state-of-the-art pure data-driven models alongside physics-aware models, spanning five two-dimensional non-trivial benchmarks. Notably, PAPM achieved an average absolute performance boost of 6.7\% with fewer FLOPs and only 1\%-10\% of the parameters compared to alternative methods.
\end{itemize}
\section{Related Work}\label{RelatedWork}
\textbf{Pure Data-driven Method.} 
There are various neural network designs, where CNNs~\citep{yu2017dilated, bhatnagar2019prediction, stachenfeld2021learned} and GNNs~\citep{sanchez2020learning, li2022graph} target spatial dynamics within mesh grids, while RNN~\citep{kochkov2021machine} and LSTM~\citep{shi2015convolutional, zhang2020physics} focus on temporal progression. 
Another line is the neural operator, excelling in mapping between temporal-spatial functional spaces, demonstrating success across various PDEs. Fourier neural operator (FNO)~\citep{li2020fourier} learns the operator by harnessing the spectral domain alongside the Fast Fourier Transform. DeepONet~\citep{lu2019deeponet} approximates various nonlinear operators by leveraging branch and trunk networks for input functions and query points. Building upon this, MIONet~\citep{jin2022mionet} addresses the challenges of multiple input functions within the DeepONet framework. Moreover, U-FNets~\citep{gupta2022towards} and convolutional neural operators (CNO)~\citep{raonic2023convolutional} are modified U-Net~\citep{ronneberger2015u} variants, where the former replace U-Net's layers by FNO's Fourier blocks, and the latter replace them by predefined convolutional block.

\textbf{Physics-aware Method.} 
Contrary to the method of integrating complete physics knowledge into its loss function, the physics-aware method only leverages labels while explicitly incorporating either entire or partial mechanistic knowledge into the network architecture. 
Inspired by the finite volume method, FINN~\citep{karlbauer2022composing} innovatively employs flux and state kernels for modeling components of advection-diffusion equations in each volume, which is one class of process systems, and has an explicit form. Since FINN is conducted for each volume, FINN is computationally inefficient in the whole spatial region. PiNDiff~\citep{akhare2023physics}, PeRCNN~\citep{rao2023encoding}, and PPNN~\citep{liu2024multi} are inspired by the finite difference method.
PiNDiff~\citep{akhare2023physics} and NeuralGCM \citep{dmitrii2023Neural} integrate partial physics knowledge into the NN block for forecasting the systems' future states, ensuring mathematical integrity via differentiable programming. 
PeRCNN~\citep{rao2023encoding} employs convolutional operations to approximate unknown nonlinear terms for reaction-diffusion systems while incorporating known terms through difference schemes.
PPNN~\citep{liu2024multi} bakes prior-knowledge terms from low-resolution data, estimates unknown parts with the trainable network, and uses the Euler time-stepping difference scheme to form a regression model for updating states.

\textbf{Learned correction methods.}
As highlighted in \citep{rackauckas2020universal, um2020solver, dresdner2022learning, sun2023neural, mcgreivy2023invariant}, the core idea of learned correction methods is to approximate the known part with specific fixed modules (such as numerical methods) and approximate the unknown part with a neural network, which often yields superior results compared to fully learned approaches. However, the current learned correction methods typically rely on known equations and are optimized for specific conditions, which is somewhat different from our model's broader objective of generalizing across various conditions and conservation relations. The second difference between our work and the mentioned literature lies in the precision of the known parts. Specifically, PAPM only knows that the different terms in the equation adhere to the general form of conservation relations, without exact knowledge of each term's specific composition. In contrast, the literature deals with cases where the known parts are precise. 

\section{Preliminaries}\label{Preliminaries}
This section presents the foundational description of process systems, known as the process model. Additionally, further clarification is provided on the specific problem in this work.

\textbf{Process Model.} 
Pivotal in engineering disciplines, process models represent and predict the dynamics of diverse process systems. This model's mathematical foundation relies on two essential sets of equations: conservation equations, governing the dynamic behavior of fundamental quantities, and constitutive equations, which describe the interactions among different variables. Further details are provided in Appendix~\ref{Process_models}.

Eq.~\ref{conservation} and Eq.~\ref{constitutive} represent the universal conservation and constitutive equations, respectively.
\vspace{-2mm}
\begin{equation}\label{conservation}
\left\{
\begin{aligned}
&\frac{\partial \boldsymbol{U}^{t}}{\partial t}=-\nabla\cdot\left(\boldsymbol{J}_{C}+\boldsymbol{J}_{D}\right)+\boldsymbol{q}+\boldsymbol{F}\\
&\boldsymbol{J}_{\boldsymbol{C}}=\boldsymbol{U}^{t} \cdot \boldsymbol{v},\quad \boldsymbol{J}_{\boldsymbol{D}}=-\boldsymbol{D} \cdot \nabla \boldsymbol{U}^{t}
\end{aligned}
\right.
\vspace{-1mm}
\end{equation}
\begin{equation}\label{constitutive}
\left\{
\begin{aligned}
&\boldsymbol{v} = v(\boldsymbol{U}^{t}),\quad \boldsymbol{D}=\boldsymbol{\lambda},\\
&\boldsymbol{q}=h_O(\boldsymbol{U}^{t}),\quad \boldsymbol{F}=h_F\left(\boldsymbol{X}_F\right)
\end{aligned}
\right.
\vspace{-1mm}
\end{equation}
where $\boldsymbol{U}$ is the physical quantity, denoting the system's state. 
Eq.~\ref{conservation} comprises four essential elements: the diffusion flows $\boldsymbol{J}_{D}$, convection flows $\boldsymbol{J}_{C}$, the internal source $\boldsymbol{q}$, and the external source $\boldsymbol{F}$. In Eq.~\ref{constitutive}, $\boldsymbol{v}$ denotes the velocity of the physical quantity being transmitted, $\boldsymbol{D}$ is the diffusion coefficient, $\boldsymbol{\lambda}$ denotes the coefficients, and $\boldsymbol{X}_F$ is the input of the external source term. Here, the corresponding linear or nonlinear mapping is the $v$, $h_O$, and $h_F$.

\begin{figure}[t]
\centering
\includegraphics[width=0.45\textwidth]{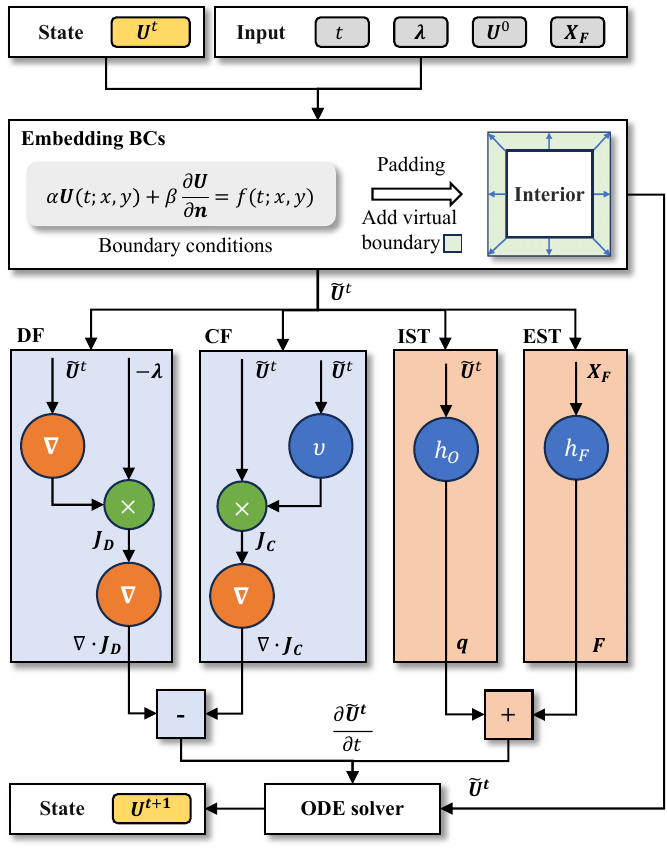}
\vspace{-2mm}
\caption{A detailed structure of PAPM at time $t$. 
Here, $v$, $h_O$, and $h_F$ are the corresponding unknown mapping, and the neural networks are needed for learning. We propose a temporal-spatial stepping module (TSSM) for DF, CF, IST, and EST in section~\ref{tssm_all}, which aligns with the distinct equation characteristics of different process systems.}
\label{PAPM}
\end{figure}

The structure of PAPM is depicted in Fig.~\ref{PAPM} at time $t$ $(\boldsymbol{U}^{t}\rightarrow\boldsymbol{U}^{t+1})$.
Our goal is to use partial prior knowledge (the general form of Eq.~\ref{conservation}) and a small amount of label data to establish a proxy model, which takes these various input conditions and outputs system time-dependent solutions ($\{\boldsymbol{U}^{t}\}_{1\leq t \leq T}$), as shown in Fig~\ref{pipline}. Notably, the general form of Eq.~\ref{conservation} is only known, which is also the data flow, while the specific item is unknown, such as the mappings of $v$, $h_O$, and $h_F$, aligning with most real-world scenarios~\citep{karlbauer2022composing, huang2023learning, liu2024multi}.

\textbf{Problem Formulation.} 
Under different initial and boundary conditions, external sources, and coefficients, the following $T$-step trajectory should be predicted. 
Moreover, due to the high cost of generating labeled data, we focused on out-of-sample scenarios, \textit{e.g.} time extrapolation, where the training dataset only contains the subsequent $T'$-step trajectory, and $1\leq T'\ll T$.

Formally, the dataset $\mathcal{D} = \{(\boldsymbol{a}_k, \boldsymbol{S}_k)\}_{1\leq k \leq D}$, where $\boldsymbol{S}_k = \mathcal{G}(\boldsymbol{a}_k)$. 
Here $\boldsymbol{a}_k$ contains a set of inputs, that is, initial condition $\boldsymbol{U}_k^0$, boundary conditions, such as Robin conditions, external sources $\boldsymbol{X}_F$, and coefficients $\boldsymbol{\lambda}$.
$\boldsymbol{S}_k = (\boldsymbol{U}_k^{1}, \cdots, \boldsymbol{U}_k^{T})$ is the following trajectory, and the mapping $\mathcal{G}(\cdot)$ is our goal to learn. 
Each $\boldsymbol{U}_k^{t} = \{\boldsymbol{u}_{k, i}^{t}\}_{1\leq i \leq m}$ is a vector, which consists of $m \in \mathbb{N}^{+}$ physical quantities, such as velocity, vorticity, pressure, and temperature.
We discretize each quantity $\boldsymbol{u}_{k, i}^{t}\in \mathcal{R}^{N}$ on the grid $\left\{\boldsymbol{x}_j \in \Omega\right\}_{1 \leq j \leq N}$. In a nutshell, for modeling this operator $\mathcal{G}(\cdot)$, we use a parameterized neural network $\hat{\mathcal{G}}_{\theta}$ with parameters $\theta$, inputs $\boldsymbol{a}_k$, and outputs $\hat{\mathcal{G}}_{\theta}(\boldsymbol{a}_k) = \hat{\boldsymbol{S}}_k$, where $1\leq k \leq D$. 

Our goal is to minimize the $L_2$ relative error loss between the prediction $\hat{\boldsymbol{S}}_k$ and real data $\boldsymbol{S}_k$ in the training dataset as,
\vspace{-4mm}
\begin{equation}
\begin{aligned}
&\min_{\boldsymbol{\theta} \in \boldsymbol{\Theta}} \frac{1}{D_0} \sum_{k=1}^{D_0} \mathcal{L}_k(\boldsymbol{\theta})
=\min_{\boldsymbol{\theta} \in \boldsymbol{\Theta}}
\frac{1}{D_0} \sum_{k=1}^{D_0}
\frac{1}{T'} \sum_{t=1}^{T'} 
\mathcal{L}_k(t,\boldsymbol{\theta})\\
&=\min_{\boldsymbol{\theta} \in \boldsymbol{\Theta}}
\frac{1}{D_0} \sum_{k=1}^{D_0}
\frac{1}{T'}\sum_{t=1}^{T'}
\frac{1}{m}\sum_{i=1}^{m}
\frac{\parallel \boldsymbol{u}_{k,i}^{t} - \hat{\boldsymbol{u}}_{k,i}^{t} \parallel_2}{\parallel \boldsymbol{u}_{k,i}^{t} \parallel_2}
\end{aligned}
\label{all_lossfunction}
\end{equation}
where $T'$ is the training time-step size, and $D_0$ is the size of the training dataset. $\mathcal{L}_k(t,\boldsymbol{\theta}) = \frac{1}{m}\sum_{i=1}^{m}
\frac{\parallel \boldsymbol{u}_{k,i}^{t} - \hat{\boldsymbol{u}}_{k,i}^{t} \parallel_2}{\parallel \boldsymbol{u}_{k,i}^{t} \parallel_2}$ is the mean $L_2$ relative error loss at time $t$ of index $k$. $\boldsymbol{\theta}$ is a set of the network parameters and $\boldsymbol{\Theta}$ is the parameter space. 

\section{Methodology}
This section presents PAPM's architecture specifically tailored to conservation and constitutive relations. Then, a holistic temporal-spatial stepping module (TSSM) is proposed, adapting to the unique equation characteristics of various process systems. 
The Appendix~\ref{Algorithm} provides the pseudo-code for the entire training process, offering a comprehensive understanding of our approach.
\subsection{PAPM Overview}\label{PAPM_Overview}
Aligning with the general form of Eq.~\ref{conservation} and Eq.~\ref{constitutive}, there are four elements corresponding to Diffusive Flows (DF), Convective Flows (CF), Internal Source Term (IST), and External Source Term (EST) in PAPM's structure diagram, as illustrated in Fig.~\ref{PAPM}.
The versatile general structure of PAPM could enable it to work effectively across different process systems.
The input contains a set of inputs, that is, coefficient $\boldsymbol{\lambda}$, initial state $\boldsymbol{U}^0$, external source input $\boldsymbol{X}_F$, and boundary conditions, which are multiple conditions of process systems.
The sequence of embedding this prior knowledge unfolds as follows:

\input{tssm_table}

\textbf{1) Embedding BCs.} Using the given boundary conditions, the physical quantity $\boldsymbol{U}^t$ is updated, yielding $\tilde{\boldsymbol{U}}^t$. A padding strategy is employed to integrate four different boundary conditions in four different directions into PAPM. Further details are provided in Appendix~\ref{embedding}.

\textbf{2) Diffusive Flows (DF).} Using $\tilde{\boldsymbol{U}}^t$ and coefficients $\boldsymbol{\lambda}$, we represent the directionless diffusive flow. The diffusion flow and its gradient are obtained as $\boldsymbol{J}_{D}=-\boldsymbol{D} \cdot \nabla \tilde{\boldsymbol{U}}^t$ and $\nabla \cdot \boldsymbol{J}_{D}$ via a symmetric gradient operator, respectively. 

\textbf{3) Convective Flows (CF).} The pattern $v$ is derived from $\tilde{\boldsymbol{U}}^t$. Once $\boldsymbol{v}$ is determined, its sign indicates the direction of the flows, enabling computation of $\boldsymbol{J}_C=\tilde{\boldsymbol{U}}^t \cdot \boldsymbol{v}$ and $\nabla \cdot \boldsymbol{J}_C$ through a directional gradient operator. 

\textbf{4) Internal Source Term (IST) \& External Source Term (EST).} Generally, IST and EST present a complex interplay between physical quantities $\tilde{\boldsymbol{U}}^t$ and external inputs $\boldsymbol{X}_F$. Often, this part in real systems doesn't have a clear physics-based relation, prompting the use of NNs to capture this intricate relationship. 

\textbf{5) ODE solver.}  From DF, CF, IST, and EST, the dynamic $\partial \tilde{\boldsymbol{U}}^t / \partial t$ are derived. 
By doing so, the Eq.~\ref{conservation} can be reduced to an ordinary differential equation (ODE), and the ODE solver is used to approximate the evolving state as $\boldsymbol{U}^{t+1}=\tilde{\boldsymbol{U}}^t+\int_t^{t+1} \frac{\partial \tilde{\boldsymbol{U}}^t}{\partial t}dt$.

Fig.~\ref{PAPM} above illustrates the data flow at time $t$. Then, during the training or inference phase, PAPM performs autoregressive predictions as $\boldsymbol{U}^{t+1} = \mathcal{G}_\theta(\boldsymbol{U}^{t}, a_k)$, where $1 \leq t \leq \tau$, with $\tau = T'$ during training and $\tau = T$ during inference. 

In short, PAPM takes different conditions, including initial conditions, boundary conditions, external sources, and coefficients, and interactively propagates the dynamics of process systems forward using five distinct components. The purpose of such a structured design is to reinforce the inductive biases concerning strict physical laws.

\subsection{Temporal-Spatial Stepping Module (TSSM)}\label{tssm_all}
Due to the diversity of process systems, we develop a holistic Temporal-Spatial Stepping Module (TSSM) to align with the unique characteristics of different process systems, which forms the specific network structure for each component in PAPM. As shown in Tab.~\ref{tssm_tab}, TSSM is categorized into three types based on structures of process systems, where each type decomposes temporal and spatial components, \textit{i.e.}, structure-preserved localized operator, spectral operator, and hybrid operator. Notably, all three approaches can employ a common temporal operation through ODE solvers.

\begin{figure}[h]
\centering
\includegraphics[width=0.48\textwidth]{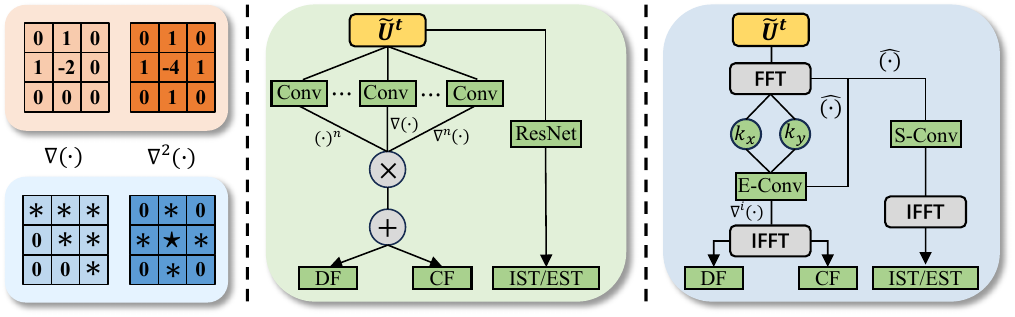}
\vspace{-6mm}
\caption{\textbf{Left:} Pre-defined convolutional kernels, where fixed and trainable correspond to the matrices at the top and bottom, respectively. The bottom kernels approximate the unidirectional convection (upwind scheme) and directionless diffusion (central scheme). Symbols $\bm{\ast}$ and $\bm{\star}$ indicate trainable parameters corresponding to the upper triangular and symmetric matrices. 
\textbf{Mid:} Structure-preserved localized operator. 
\textbf{Right:} Structure-preserved spatial operator.}
\label{TSSM}
\end{figure}

\textbf{Temporal Operation.} After obtaining the dynamic state derivative, $\partial \tilde{\boldsymbol{U}}^t/\partial t$, the subsequent state $\boldsymbol{U}^{t+1}$ can be computed through numerical integration over different time spans. Due to the numerical instability associated with first-order explicit methods like the Euler method~\citep{gottlieb2001strong, fatunla2014numerical}, we adopt the neural ordinary differential equation approach (Neural ODE~\citep{chen2018neural}), which employs the Runge–Kutta time-stepping strategy to enhance stability. The computed state $\boldsymbol{U}^{t+1}$ is then recursively fed back into the network as the input for the subsequent time step, continuing this process until the final time step is reached.

\textbf{Structure-preserved Localized Operator.} 
For systems with explicit structures, such as the Burgers and RD equations, typified by expressions like $-u\nabla u + \nabla^2 u$, convolutional kernels in the physical space are employed to capture system dynamics. 
We opt for either fixed or trainable kernels, illustrated in Fig.~\ref{TSSM} (Left), depending on our understanding of the system. 
Specifically, the fixed one is based on the predefined convolution kernel derived from difference schemes, and further details are provided in Appendix~\ref{localized operator}. 
Moreover, the trainable version tailors its design to essential features of convection (upper or lower triangular) and diffusion (symmetric). 
Once set, the localized operator is depicted in Fig.~\ref{TSSM} (Mid). We could represent nonlinear terms in DF and CF using predefined convolution kernels alongside partially known physics (the general form of Eq.~\ref{conservation} and Eq.~\ref{constitutive}).
Any unknown specific terms, such as source, are then addressed through a shallow ResNet block.

\textbf{Structure-preserved Spectral Operator.} 
For systems with implicit structures, such as the Navier-Stokes Equation in vorticity form, represented like $-u\nabla w+\nabla^2 w$, we adopt a sequential process, as shown in Fig.~\ref{TSSM}(Right). 
Recognizing the implicit linkage between velocity and vorticity, $w$ is initially processed to extract the flow function, subsequently leading to the velocity derivation. 
The spectral space dimensions ($k_x$ and $k_y$) and spectral quantity (denoted as $\hat{\cdot}$), are obtained by leveraging the FFT. 
Associating $k_x$ and $k_y$ with $\hat{\cdot}$, differential operators like $\nabla^{i}(\cdot)$ are represented via E-Conv (\textit{e.g.}, element-wise product), and then mapped back to the physical space using IFFT. Doing so can represent the nonlinear terms in DF and CF via simple computations such as addition and multiplication in the spectral domain.
Moreover, the spectral convolution (S-Conv) fo FNO~\citep{li2020fourier} is introduced to learn unknown components. This process can be further detailed in Appendix~\ref{spectral operator}. 

\textbf{Structure-preserved Hybrid Operator.} For systems with a hybrid structure, such as the Navier-Stokes Equation in general form (\textit{e.g.}, $-u\nabla u + \nabla^2 u - \nabla p$), given the implicit interrelation between pressure $p$ and velocity $u$, a combination of the method above is employed. Explicit constituents, such as $u\nabla u$ and $\nabla^2 u$, are addressed through the localized operator. Meanwhile, implicit relations are resolved similarly by the spectral operator. For unknown components, either of the two operators can be engaged. We generally favor the localized operator as it allows direct operations without requiring transitions between different spaces.

\section{Experiments}\label{experiments}
\subsection{Experimental setup and evaluation protocol}
\textbf{Datasets.} We employ five datasets spanning diverse domains, such as fluid dynamics and heat conduction, detailed in Appendix~\ref{data}. These datasets are categorized based on the Temporal-Spatial Stepping Module used in PAPM to highlight distinct equation characteristics in various process systems.
\vspace{-2mm}
\begin{itemize}
    \vspace{-1mm}
    \item \textbf{Localized Category:} 
    \textbf{Burgers2d~\citep{huang2023learning}} is a 2D benchmark PDE with periodic boundary conditions, various initial conditions, and viscosity, while the source remains unknown.
    \textbf{RD2d~\citep{takamoto2022pdebench}} addresses a 2D FitzHugh-Nagumo reaction-diffusion equation with no-flow Neumann boundary condition, diverse initial conditions, and unknown source terms.
    \vspace{-1mm}
    \item \textbf{Spectral Category:} 
    \textbf{NS2d~\citep{li2020fourier}} explores incompressible fluid dynamics in vorticity form with varied initial conditions and unknown sources.
    \vspace{-1mm}
    \item \textbf{Hybrid Category:} 
    \textbf{Lid2d} focuses on incompressible lid-driven cavity flow, characterized by differing viscosity and BCs. \textbf{NSM2d} deals with incompressible fluid dynamics within an unknown magnetic field source, featuring time-varying BCs, various initial conditions, and viscosity.
\end{itemize}
\vspace{-2mm}

\textbf{Baselines.} We compared our approach with eight SOTA baselines for a comprehensive evaluation. \textbf{ConvLSTM}~\citep{shi2015convolutional} is a classical time series modeling technique that captures dynamics via CNN and LSTM. 
\textbf{Dil-ResNet}~\citep{stachenfeld2021learned} adopts the encoder-process-decoder process with dilated-ConvResNet for dynamic data through an autoregressive stepping manner. 
\textbf{time-FNO2D}~\citep{li2020fourier} and \textbf{MIONet}~\citep{jin2022mionet} are two typical neural operators in learning dynamics. 
\textbf{U-FNet}~\citep{gupta2022towards} and \textbf{CNO}~\citep{raonic2023convolutional} are modified U-Net~\citep{ronneberger2015u} variants.
\textbf{PeRCNN}~\citep{rao2023encoding} incorporates specific physical structures into a neural network, ideal for sparse data scenarios. \textbf{PPNN}~\citep{liu2024multi} is a novel autoregressive framework preserving known PDEs using multi-resolution convolutional blocks.

\textbf{Metrics.} 
According to Eq.~\ref{all_lossfunction}, we adopt the mean $L_2$ relative error, abbreviated as $\epsilon$, as our evaluation metric for validation and testing datasets. $\epsilon$ is formulated as follows:
\begin{equation}
\epsilon=
\frac{1}{D} \sum_{k=1}^{D}
\frac{1}{T}\sum_{t=1}^{T}
\frac{1}{m}\sum_{i=1}^{m}
\frac{\parallel \boldsymbol{u}_{k,i}^{t} - \hat{\boldsymbol{u}}_{k,i}^{t} \parallel_2}{\parallel \boldsymbol{u}_{k,i}^{t} \parallel_2}
\end{equation}\label{all_Metrics}

\textbf{Evaluation Protocol.} 
Based on the experimental setting of time extrapolation, we further conducted experiments in the following two parts: coefficient interpolation (referred to as \textbf{C Int.}) and coefficient extrapolation (referred to as \textbf{C Ext.}). 
More information about the evaluation protocol, the hyper-parameters of baselines, and our methods can be further detailed in Appendix~\ref{all_details}. 

\input{table_1_all}

\subsection{Main Results}
\textbf{Performance Comparisons.} Tab.~\ref{main_results} and Tab.~\ref{parameters} present the primary experimental outcomes, the number of trainable parameters ($N_{P}$), and computational cost (FLOPs) for each baseline across datasets.

Here, \textbf{Bold} and \underline{Underline} indicate the best and second best performance, respectively. Notably, lower results mean better performance because the metric is the mean $L_2$ relative error. Our observations from the results are as follows:

\textbf{Firstly}, PAPM exhibits the most balanced trade-off between performance, parameter count, and computational cost among all methods evaluated, from explicit structures (Burgers2d, RD2d) to implicit (NS2d) and more complex hybrid structures (Lid2d, NSM2d). 
Remarkably, even though PAPM requires significantly fewer FLOPs and only 1\% of the parameters employed by the prior leading method, PPNN, it still outperforms it by a large margin. In a nutshell, our model enhances the performance by an average of 6.7\% over nine tasks, affirming PAPM as a versatile and efficient framework suitable for diverse process systems.

\textbf{Secondly}, PAPM's structured treatment of system inputs and states leads to a remarkable 9.6\% performance boost in three coefficient-extrapolation tasks. This highlights its superior generalization capability in out-of-sample scenarios. Unlike models like PPNN, which directly use system-specific inputs, PAPM integrates coefficient data more intricately within conservation and constitutive relations, boosting its adaptability to varying coefficients.

\textbf{Thirdly}, data-driven methods are less effective than physics-aware methods like PPNN and our PAPM in time extrapolation tasks, where incorporating prior physical knowledge through structured network design enhances a model's generalization ability. Notably, PeRCNN uses $1\times1$ convolution to approximate nonlinear terms, but experimental results suggest limited performance. Further details are available in Appendix~\ref{details}.

\begin{figure}[h]
\centering
\includegraphics[width=0.45\textwidth]{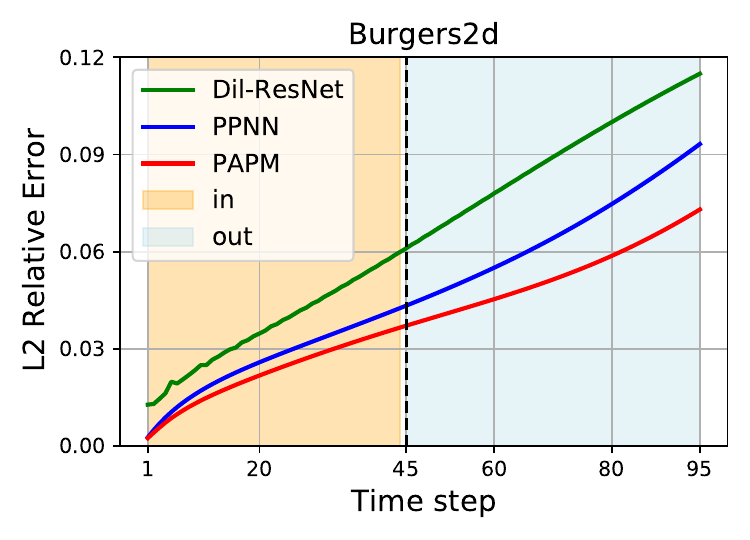}
\vspace{-6mm}
\caption{$\epsilon$ (Eq. 4) of predicted each time step on Burgers2d, where \textbf{in} is the same as the training, \textbf{out} is the time extrapolation.}\label{each_step}
\end{figure}

\begin{figure*}[t]
\centering
\includegraphics[width=\textwidth]{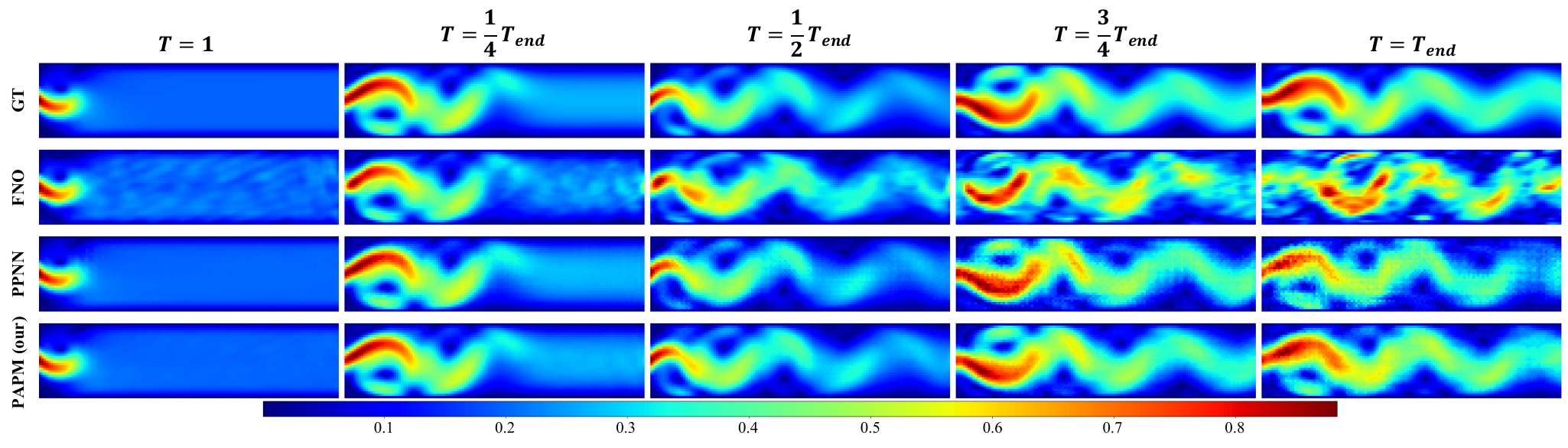}
\vspace{-2mm}
\caption{Predicted flow velocity ($|\boldsymbol{u}|_2$) snapshots by FNO, PPNN, and PAPM (Ours) vs. Ground Truth (GT) on NSM2d dataset in T Ext. task.}
\label{nsm_velocity}
\end{figure*}
\textbf{Visualization.} Fig.~\ref{each_step} showcases the stepwise relative error of PAPM during the extrapolation process in the test dataset, using Burgers2d's \text{C Int.} as a representative example.
Compared to the two best-performing baselines, our model (depicted by the red line) exhibits superior performance throughout the extrapolation process, with the least error accumulation. Turning our attention to the more challenging NSM2d dataset, Fig.~\ref{nsm_velocity} presents the results across five extrapolation time slices. While FNO demonstrates commendable accuracy within the training domain ($T\leq\frac{1}{2}T_{end}$), its performance falters significantly outside of it ($\frac{1}{2}T_{end}<T \leq T_{end}$). On the other hand, physics-aware methods (PPNN), and PAPM in particular, consistently capture the evolving patterns with a greater degree of robustness. Notably, our method emerges as a leader in terms of precision. Additional visual results can be found in Appendix~\ref{visual}.

\begin{figure*}[h]
\centering
\includegraphics[width=0.9\textwidth]{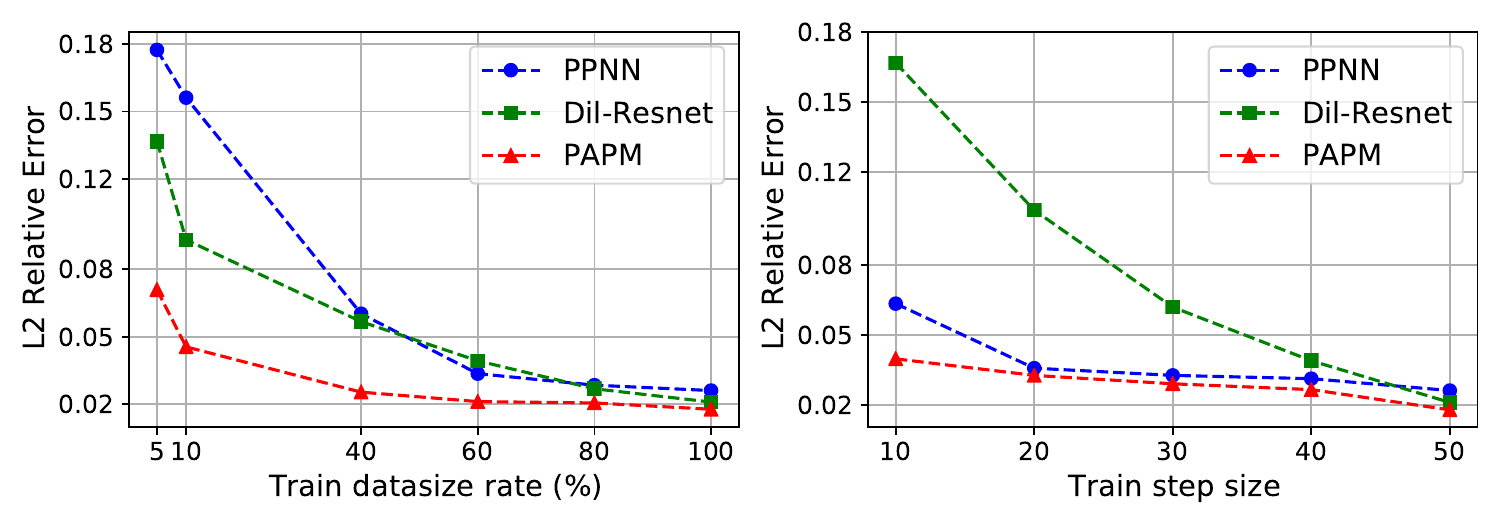}
\vspace{-2mm}
\caption{$\epsilon$ (Eq. 4) comparison by the leading method, PPNN, Dil-Resnet, PAPM (Ours) on the RD2d dataset. Left: varying amount of training data. Right: varying train step size.}\label{compare_two}
\vspace{-1mm}
\end{figure*}

\subsection{Efficiency}
\textbf{Training and Inference Cost}: Dataset generation for our work is notably resource-intensive, with inference times ranging from $10^2$ to $10^4$ seconds for public datasets, and up to $10^3$ seconds for those datasets we generated using COMSOL Multiphysics.
In stark contrast, both baselines and PAPM register inference times between $0.1$ to $10$ seconds (detailed in Appendix~\ref{cost}), achieving an improvement of 3 to 5 orders of magnitude. Notably, PAPM's time cost rivals or even surpasses baselines across different datasets. PAPM's efficiency remains competitive with other data-driven methods.

\textbf{Data Efficiency.} Owing to PAPM's structured design, data utilization is significantly enhanced. To evaluate data efficiency, we conducted tests using RD2d as a representative example, with Dil-ResNet and PPNN symbolizing pure data-driven and physics-aware methods. The results, displayed in Fig.~\ref{compare_two}, depict PAPM's efficiency concerning data volume and label data step size in training.

\textbf{(1) Amount of Data}: With a fixed 20\% reserved for the test set, the remaining 80\% of the total data is allocated to the training set. We systematically varied the training data volume, ranging from initially utilizing only 5\% of the training set and progressively increasing it to the entire 100\%.
PAPM's relative error distinctly outperforms other baselines, especially with limited data (5\%). As depicted in Fig.~\ref{compare_two} (Left), PAPM's error consistently surpasses other methods, stabilizing below 2\% as the training data volume increases. 

\textbf{(2) Time Step Size}: We varied the time step size from a tenth to half of the total duration, increasing it in increments of tenths. As shown in Fig.~\ref{compare_two} (Right), PAPM demonstrates the capability for long-range time extrapolation with fewer dynamic steps. It consistently outperforms other methods, achieving superior results even with shorter training time step sizes.

\subsection{Ablation Studies} 

\textbf{Different blocks impacts.}
Tab.~\ref{ablation_studies} displays our selection of the Burger2d dataset for ablation studies, chosen for its representation of diffusion, convection, and source terms. We defined several configurations to assess the impact of individual components. The $L_2$ relative error on the boundary (BC $\epsilon$) is introduced to highlight the importance of physics embedding further. \textbf{no\_DF} excludes diffusion, whereas \textbf{no\_CF} omits convection. In \textbf{no\_Phy}, we retain only a structure with a residual connection, thereby eliminating both diffusion and convection. \textbf{no\_BCs} setup removes the explicit embedding of boundary conditions, \textbf{no\_NODE} replaces the Neural ODE with the Euler stepping scheme, and \textbf{no\_All} adopts a purely data-driven approach. Additional ablation results can be found in Appendix~\ref{supp_ablation}.

Key findings include the crucial roles of diffusion and convection in representing system dynamics, as evidenced in the no\_DF, no\_CF, and no\_Phy configurations. Specifically, the no\_DF configuration demonstrated the importance of integrating the viscosity coefficient with the diffusion term, with its absence leading to significant errors. The necessity of adhering to physical laws in boundary conditions was highlighted in the no\_BCs, notably reducing errors on the boundary (BC $\epsilon$). Lastly, the no\_NODE results indicate that different temporal stepping schemes significantly impact the outcomes, underscoring the effectiveness of neural ODEs in continuous-time modeling.

\begin{table}[t]
\centering
\renewcommand{\arraystretch}{1}
\setlength{\tabcolsep}{8pt}
\caption{Ablation comparison of $\epsilon$ and $\epsilon$ on the boundary (BC $\epsilon$).}
\vspace{2mm}
\label{ablation_studies}
\small{
\begin{tabular}{l|cc|cc}
    \toprule
    \multirow{2}{*}{\textbf{Config}} &\multicolumn{2}{c|}{\textbf{C Int.}}& \multicolumn{2}{c}{\textbf{C Ext.}} \\ 
    &$\epsilon$ & BC $\epsilon$ & $\epsilon$ & BC $\epsilon$\\ 
    \midrule
    no\_DF         &0.067&0.051&0.207&0.067\\
    no\_CF         &0.062&0.043&0.131&0.054\\
    no\_Phy        &0.149&0.051&0.210&0.144\\ 
    no\_BCs        &0.068&0.097&0.136&0.193\\
    no\_NODE       &0.053&0.041&0.150&0.045\\
    no\_All        &0.162&0.195&0.216&0.250\\
    \midrule
    \textbf{PAPM}&\textbf{0.039}&\textbf{0.037}&\textbf{0.101}&\textbf{0.043}\\ 
    \bottomrule
\end{tabular}
}
\end{table}

\textbf{Different blocks validations.}
Taking the Burgers2d and RD2d datasets as examples to demonstrate the fact that the convection/diffusion/source terms could actually learn those parts in the equations. We use high-fidelity FDM and FVM to compute the corresponding terms to obtain the detailed term for convection/diffusion/source terms. 
The $L_2$ relative error between ground truth and numerical results are $0.0041$ and $0.0032$ in all time steps for Burgers2d and RD2d, respectively. Thus, we can use the results obtained by the numerical methods as reference values to verify this.
As shown in Tab. \ref{table:epsilon_comparison}, the different terms of PAPM can be used to learn the equation's convection/diffusion/source parts. Additional visualization results can be found in Appendix~\ref{visual} (Fig. \ref{bur_ab} and Fig. \ref{rd_ab}).

\begin{table}[h]
\centering
\renewcommand{\arraystretch}{1}
\setlength{\tabcolsep}{4pt}
\caption{Comparison of $\epsilon$ for different blocks on different datasets.}
\vspace{2mm}
\small{
\begin{tabular}{c|c|c|c|c}
\toprule
\textbf{Datasets} & $\epsilon$ & convection $\epsilon$ & diffusion $\epsilon$ & source $\epsilon$ \\ 
\midrule
Burgers2d & \textbf{0.039} & 0.037 & 0.041 & 0.069 \\ 
RD2d & \textbf{0.018} & - & 0.025 & 0.012 \\ 
\bottomrule
\end{tabular}
}
\label{table:epsilon_comparison}
\end{table}

\section{Conclusion}\label{conclusion}
To address the challenges of physics-aware models falling short regarding exploration depth and universality, we have proposed PAPM. 
It fully incorporates partial prior physics of process systems, which includes multiple input conditions and the general form of conservation relations, resulting in better training efficiency and out-of-sample generalization. 
Additionally, PAPM contains a holistic temporal-spatial stepping module for flexible adaptation across various process systems.
The efficacy of PAPM's structured design and holistic module was extensively validated across five datasets within distinct out-of-sample tasks. 
Notably, PAPM achieved an average performance boost of 6.7\% with fewer FLOPs and only 1\% of the parameters employed by the prior leading method.
Through such analysis, PAPM exhibits the most balanced trade-off between accuracy and computational efficiency among all evaluated methods, alongside impressive out-of-sample generalization capabilities.

\section{Limitation and Future Work}\label{conclusion}
We aim to extend our model to more complex, industrially relevant systems, moving beyond 2D spatio-temporal dynamics to scenarios like 3D-industry-standard aerodynamics, plasma discharge, and multi-physics couplings (e.g., fluid-structure and thermal fluid-structure interactions). 
Despite our model's proven balance in accuracy and efficiency, we aim to challenge it further in these intricate environments. Additionally, we plan to adapt PAPM for irregular grid scenarios, typical in the industry, by integrating graph neural networks. This will enhance PAPM's versatility, allowing it to handle diverse data structures and complex dynamic processes such as convection, diffusion, and source interactions.

\section*{Impact Statement} This paper presents work whose goal is to advance the field of Machine Learning. High performance requires substantial resources, including computing power and time, to generate quality data. Acquiring such data can be costly or resource-intensive in real-world systems, leading to potential resource wastage. The ethical use of these resources, avoiding unnecessary environmental impact, and ensuring the cost-effectiveness of data acquisition are critical concerns. Therefore, we emphasize the importance of efficient and responsible data usage to minimize adverse societal and environmental effects.

\section*{Acknowledgment}
This work was supported by the National Key Research and Development Program of China (No.2021YFF0500403).

\bibliographystyle{icml2024}
\bibliography{mybib}

\newpage
\appendix
\input{notations}
\onecolumn
\icmltitle{Appendix of PAPM}
\icmlsetsymbol{equal}{*}
In this appendix, we first summarize the notations in Tab.\ref{tab_notations} (\ref{all_notations}). 
Then, we describe the process model further, showing in detail the starting point of our problem (\ref{Process_models}). 
Secondly, we perform a further theoretical presentation on the details of embedding boundary conditions (\ref{embedding}).
The algorithm display and details of the proposed temporal and spatial stepping module (TSSM) are further elaborated (\ref{Algorithm}) and (\ref{operator}), respectively.
Moreover, the instructions of TSSM are discussed in (\ref{Instructions_TSSM}).
Subsequently, the five datasets are further described (\ref{data}). 
The hyper-parameters settings of different baselines and PAPM are shown in detail (\ref{all_details}).
Finally, some additional experimental results are shown (\ref{additional_results}), which are data visualization (\ref{visual}), training and inference time cost-specific details (\ref{cost}), and detailed ablation studies (\ref{supp_ablation}).

\section{Table of notations}\label{all_notations}
A table of notations is given in Tab.\ref{tab_notations}.

\section{Process models}\label{Process_models}
Pivotal in engineering disciplines, process models serve to represent and predict the dynamics of diverse process systems, from entire plants to single equipment pieces. These models primarily rely on the interplay between conservation and constitutive equations, ensuring an accurate depiction of system dynamics. Conservation equations dictate the model's dynamics using partial differential equations that govern primary physical quantities like mass, energy, and momentum. On the other hand, constitutive equations relate potentials to extensive variables via algebraic equations, such as flows, temperatures, pressures, concentrations, and enthalpies, enriching the model's comprehensiveness. Additionally, accounting for initial and boundary conditions ensures the model's reliability, making these four components interdependently integral to the model's solid mathematical framework.

\textbf{Conservation equations.} The general form in differential representation is:
\begin{equation}
\begin{aligned}
\frac{\partial \boldsymbol{U}}{\partial t}=-\nabla\cdot\left(\boldsymbol{J}_{C}+\boldsymbol{J}_{D}\right)+\boldsymbol{q}+\boldsymbol{F}
\end{aligned}
\end{equation}
where $\boldsymbol{J}_{C}=\boldsymbol{U} \cdot \boldsymbol{v}$, $\boldsymbol{J}_{D}=-\boldsymbol{D} \cdot \nabla \boldsymbol{U}$, and $\boldsymbol{U}(\boldsymbol{x}, t)$ represents the state of the model, which is the object of our modeling, $\boldsymbol{x}\in \Omega$ and $t\in[0, T]$, $T\in\mathbb{R}_{+}$. $\boldsymbol{J}_{C}$ represents convective flows, and $\boldsymbol{J}_{D}$ represents diffusive or molecular flows. $\boldsymbol{v}$ describes the convective flow pattern into or out of the system volume. $\boldsymbol{D}$ represents the diffusion coefficient. $\boldsymbol{q}$, the internal source term, for example, the chemical reaction for component mass conservation, where species appear or are consumed due to reactions within the space of interest. Other internal source terms arise from energy dissipation, conversion, compressibility, or density changes. $\boldsymbol{F}$, the external source term, including gravitational, electrical, and magnetic fields as well as pressure fields.

\textbf{Constitutive equations.} For the internal source term, it usually depends on the state of the model and can be expressed as $\boldsymbol{q} = h_{q}(\boldsymbol{U}, \boldsymbol{x}, t)$. The external source term is usually related to the external effects imposed and can be expressed as $\boldsymbol{F} = h_{F}(\boldsymbol{X}_{F})$, where $\boldsymbol{X}_{F}$ is a vector of parameters imposed externally, which may include voltages, pressures, etc. For convective flows, the velocity $\boldsymbol{v}$ may be determined by the state of the model, which can be expressed as $\boldsymbol{v} = g(\boldsymbol{U}, \boldsymbol{x}, t)$. 

\textbf{Initial conditions (IC).} Every process or system evolves over time, but we need a reference or a starting point to predict or understand this evolution. The initial conditions provide this starting point. For example, in the context of a reactor, initial conditions might describe the concentration of various reactants at $ t = 0$. Mathematically, IC can be represented as $\boldsymbol{U}(\boldsymbol{x}, 0) = \boldsymbol{U}_0(\boldsymbol{x})$.

\textbf{Boundary conditions (BC).} Initial conditions set the foundation at $t=0$, while boundary conditions inform how a system evolves and interacts with its environment, for instance, by specifying heat flux at a heat exchanger's boundary or flow rate at a reactor's inlet. These boundary conditions can be categorized as Dirichlet, prescribing specific values like temperature on the boundary; Neumann, defining derivatives or fluxes such as the heat flux; and Robin, which combines aspects of both Dirichlet and Neumann, encompassing parameters like both heat transfer rates and surface temperatures. Regardless of the type, they're mathematically expressed as $\boldsymbol{U}(\boldsymbol{x}_b, t) = \boldsymbol{f}_b(t)$, where $\boldsymbol{x}_b \in \partial \Omega$.
\section{Embedding Boundary Conditions}\label{embedding}
This part covers the method of embedding four different boundary conditions (\textbf{Dirichlet}, \textbf{Neumann}, \textbf{Robin}, and \textbf{Periodic}) into neural networks via convolution padding. Let's consider a rectangular region in a $2D$ space, $\Omega=[0,a]\times [0,b]$, which can be discretized into an $M\times N$ grid, $\delta x = \frac{a}{M}$, $\delta y = \frac{b}{N}$. Each grid point can be represented as $X_{ij}=(x_i, y_j)$, where $i=1,2,...,M$ and $j=1,2,...,N$. Hence, we can transform the continuous space into a discrete grid of points.

\textbf{Boundary Conditions on the $X$-axis}. The direction vector is $\mathbf{n} = (1, 0)^T$, which means the boundary conditions are the same for each $y$ value.
\begin{itemize}
    \item \textbf{Dirichlet}: If the boundary condition is given as $\boldsymbol{U}(X,t)=f(X,t), X\in \partial \Omega$, the discrete form would be $\boldsymbol{U}_{Mj}=f_{j}$, and we can use a padding method in the convolution kernel $\boldsymbol{U}_{Mj}=f_{j}$.
    \item \textbf{Neumann}: If the boundary condition is given as $\frac{\partial \boldsymbol{U}(X,t) }{\partial \mathbf{n}} = f(X,t), X\in \partial \Omega$, the discrete form would be $\frac{\boldsymbol{U}_{(M+1)j} -\boldsymbol{U}_{(M-1)j}}{2\delta x}=f_{j}$ and we can use a padding method in the convolution kernel $\boldsymbol{U}_{(M+1)j}=\boldsymbol{U}_{(M-1)j} + (2\times \delta x)\times f_{j}$.
    \item \textbf{Robin}: If the boundary condition is given as $\alpha \boldsymbol{U}(X,t) + \beta \frac{\partial \boldsymbol{U}(X,t) }{\partial \mathbf{n}} = f(X,t), X\in \partial \Omega$, the discrete form would be $\alpha \boldsymbol{U}_{Mj} + \beta \frac{\boldsymbol{U}_{(M+1)j} -\boldsymbol{U}_{(M-1)j}}{2\times \delta x}=f_{j}$. We can use a padding method in the convolution kernel $\boldsymbol{U}_{(M+1)j}=\frac{2\times \delta x}{\beta} (f_{j} - \alpha \boldsymbol{U}_{Mj}) + \boldsymbol{U}_{(M-1)j}$.
    \item \textbf{Periodic}: If the boundary condition is given as $\boldsymbol{U}(X_1,t)=\boldsymbol{U}(X_2,t), X_1 \in \partial \Omega_1, X_2 \in \partial \Omega_2$, where $\Omega_1$ denotes the left boundary and $\Omega_2$ the right boundary, the discrete form would be $\boldsymbol{U}_{Mj}=\boldsymbol{U}_{1j}$. We can use a padding method in the convolution kernel $\boldsymbol{U}_{Mj}=\boldsymbol{U}_{1j}, \boldsymbol{U}_{(M+1)j}=\boldsymbol{U}_{2j}$.
\end{itemize}

\textbf{Boundary Conditions on the $Y$-axis}. The direction vector is $\mathbf{n} = (0,1)^T$. The basic handling method is similar to the $x$-direction case but with the grid spacing replaced with $\delta y$, and the boundary conditions applied to the upper and lower boundaries, \textit{i.e.}, $j=1$ and $j=N$. The corresponding $y$-direction expressions can be derived by replacing $x$ with $y$ in the $x$-direction expressions and swapping $i$ with $j$.

\textbf{Arbitrary Direction Boundary Conditions in the Rectangular Area.} The direction vector $\mathbf{n} = (cos(\theta), sin(\theta))^T$. Both $x$ and $y$ directions need to be considered, resulting in the following expressions for each of the four boundary conditions:
\begin{itemize}
    \item \textbf{Dirichlet}:Given the condition $\boldsymbol{U}(X,t)=f(X,t)$, its discrete form remains $\boldsymbol{U}_{ij}=f_{ij}$. The corresponding padding method in the convolution kernel is $\boldsymbol{U}_{ij}=f_{ij}$.
    \item \textbf{Neumann}: For the boundary condition $\frac{\partial \boldsymbol{U}(X,t)}{\partial \mathbf{n}}=f(X,t)$, the discrete form can be represented as $cos(\theta) \frac{\boldsymbol{U}_{(i+1)j} - \boldsymbol{U}_{(i-1)j}}{2 \delta x} + sin(\theta) \frac{\boldsymbol{U}_{i(j+1)} - \boldsymbol{U}_{i(j-1)}}{2 \delta y}=f_{ij}$. The corresponding padding method in the convolution kernel can be written as $\boldsymbol{U}_{(i+1)j} = \boldsymbol{U}_{(i-1)j} + 2 cos(\theta) \delta x f_{ij}$ and $\boldsymbol{U}_{i(j+1)} = \boldsymbol{U}_{i(j-1)} + 2 sin(\theta) \delta y f_{ij}$.
    \item \textbf{Robin}: Given the condition $\alpha \boldsymbol{U}(X,t) + \beta \frac{\partial \boldsymbol{U}(X,t)}{\partial \mathbf{n}}=f(X,t)$, the discrete form becomes $\alpha \boldsymbol{U}_{ij} + \beta cos(\theta) \frac{\boldsymbol{U}_{(i+1)j} - \boldsymbol{U}_{(i-1)j}}{2 \delta x} + \beta sin(\theta) \frac{\boldsymbol{U}_{i(j+1)} - \boldsymbol{U}_{i(j-1)}}{2 \delta y}=f_{ij}$. The corresponding padding method in the convolution kernel is $\boldsymbol{U}_{(i+1)j} = \frac{1}{\beta cos(\theta)} [f_{ij} - \alpha \boldsymbol{U}_{ij}] \times 2 \delta x + \boldsymbol{U}_{(i-1)j}$ and $\boldsymbol{U}_{i(j+1)} = \frac{1}{\beta sin(\theta)} [f_{ij} - \alpha \boldsymbol{U}_{ij}] \times 2 \delta y + \boldsymbol{U}_{i(j-1)}$.
    \item \textbf{Periodic}: For the condition $\boldsymbol{U}(X_1,t)=\boldsymbol{U}(X_2,t)$, the discrete form is $\boldsymbol{U}_{Mj}=\boldsymbol{U}_{1j}$ and $\boldsymbol{U}_{Ni}=\boldsymbol{U}_{1i}$. The corresponding padding method in the convolution kernel is $\boldsymbol{U}_{Mj}=\boldsymbol{U}_{1j}, \boldsymbol{U}_{(M+1)j}=\boldsymbol{U}_{2j}$ and $\boldsymbol{U}_{Ni}=\boldsymbol{U}_{1i}, \boldsymbol{U}_{i(N+1)}=\boldsymbol{U}_{i2}$.
\end{itemize}

\textbf{Directionless Boundary Conditions in the Rectangular Area.} The following strategies are employed for handling the Neumann and Robin boundary conditions:
\begin{itemize}
    \item For $\frac{\partial \boldsymbol{U}}{\partial X} = f(X,t)$, the discrete form is $\frac{\boldsymbol{U}_{(i+1)j} - \boldsymbol{U}_{(i-1)j}}{2 \delta x} = f_{ij}$ and $\frac{\boldsymbol{U}_{i(j+1)} - \boldsymbol{U}_{i(j-1)}}{2 \delta y} = f_{ij}$. We can use a padding method in the convolution kernel where $\boldsymbol{U}_{(i+1)j} = \boldsymbol{U}_{(i-1)j} + 2 \delta x f_{ij}$ and $\boldsymbol{U}_{i(j+1)} = \boldsymbol{U}_{i(j-1)} + 2 \delta y f_{ij}$.
    \item For $\alpha \boldsymbol{U}(X,t) + \beta \frac{\partial \boldsymbol{U}(X,t)}{\partial X} = f(X,t)$, the discrete form is $\alpha \boldsymbol{U}_{ij} + \beta \frac{\boldsymbol{U}_{(i+1)j} - \boldsymbol{U}_{(i-1)j}}{2 \delta x} = f_{ij}$ and $\alpha \boldsymbol{U}_{ij} + \beta \frac{\boldsymbol{U}_{i(j+1)} - \boldsymbol{U}_{i(j-1)}}{2 \delta y} = f_{ij}$. We can use a padding method in the convolution kernel where $\boldsymbol{U}_{(i+1)j} = \frac{1}{\beta} [f_{ij} - \alpha \boldsymbol{U}_{ij}] \times 2 \delta x + \boldsymbol{U}_{(i-1)j}$ and $\boldsymbol{U}_{i(j+1)} = \frac{1}{\beta} [f_{ij} - \alpha \boldsymbol{U}_{ij}] \times 2 \delta y + \boldsymbol{U}_{i(j-1)}$.
\end{itemize}

\section{Supplemental TSSM}
\subsection{Algorithm Display}\label{Algorithm}
Here, we provide the pseudo-code for the Temporal-Spatial Stepping Module (TSSM) training, offering a comprehensive understanding of our approach. As shown in Alg.~\ref{alg1_localized}, the structure-preserved localized operator is detailed. The latter is shown in Alg.~\ref{alg2_spectral}, and the third one, the hybrid operator, is a combination of these two operators. 

\vspace{-4mm}
\begin{minipage}[t]{0.48\textwidth}
\input{alg1}

\end{minipage}
\hfill
\begin{minipage}[t]{0.48\textwidth}
\input{alg2}
\end{minipage}
\subsection{Details of TSSM}\label{operator}
\subsubsection{Structure-preserved localized operator}\label{localized operator}
\textbf{Fixed convolution operations.} The differential operator can be approximated via convolution operations. For a one-dimensional function $u(x)$, we could use a convolution kernel of the form:
\begin{equation}
K = \frac{1}{2 \Delta x}[-1, 0, 1]
\end{equation}
where $\Delta x$ represents the step size. This convolution operation, corresponding to this kernel, can approximate the first-order central difference operator as follows:
\begin{align}
u^{\prime}(x) \approx \frac{u(x+\Delta x)-u(x-\Delta x)}{2 \Delta x} \approx u(x) \circledast K,    
\end{align}
with $\circledast$ denoting the convolution operation. For a two-dimensional function, it can be decomposed into a convolution of two one-dimensional functions. Assuming $u(x, y)$ is a two-dimensional function, the kernel could be formed as:
\begin{equation}
\begin{aligned}
& K=\frac{1}{h^2}\left[\begin{array}{ccc}
0 & 1 & 0 \\
1&-4&1 \\
0 & 1 & 0
\end{array}\right]
=\frac{1}{h^2}\left[\begin{array}{ccc}
0 & 0 & 0 \\
1&-2&1 \\
0 & 0 & 0
\end{array}\right]+\frac{1}{h^2}\left[\begin{array}{ccc}
0 & 1 & 0 \\
0 & -2&0 \\
0 & 1 & 0
\end{array}\right],
\end{aligned}
\end{equation}
where $h=\Delta x=\Delta y$ signifies the step size. The convolution operation corresponding to this kernel can approximate the second-order central difference operator, which is:
\begin{equation}
\begin{aligned}
&\nabla^2 u(x, y)  =\frac{\partial^2 u(x, y)}{\partial x^2}+\frac{\partial^2 u(x, y)}{\partial y^2} \\
& \approx \frac{u(x+h, y)-2 u(x, y)+u(x-h, y)}{h^2}+\frac{u(x, y+h)-2 u(x, y)+u(x, y-h)}{h^2} \\
& =\frac{u(x+h, y)+u(x, y+h)-4 u(x, y)+u(x-h, y)+u(x, y-h)}{h^2} \\
& = u(x, y) \circledast K .
\end{aligned}   
\end{equation}
Analogously, different convolution kernels can approximate other orders' differential operators. Utilizing convolution operations to approximate differential operators can boost computational efficiency. Nevertheless, careful consideration is needed when choosing a convolution kernel, as different kernels can influence the stability and accuracy of the numerical solution.

\textbf{Selection of FD kernels.} FD kernels are used to approximate derivative terms in PDEs, which directly affect the computational efficiency and the reconstruction accuracy. Therefore, it is crucial to choose appropriate FD kernels for discretized-based learning frameworks. For spatio-temporal systems, we need to consider both temporal and spatial derivatives. In specific, the second-order central difference is utilized for calculating temporal derivatives, \textit{i.e.},
\begin{equation}
\frac{\partial u}{\partial t}=\frac{-u(t-\delta t, \xi, \eta)+u(t+\delta t, \xi, \eta)}{2 \delta t}+\mathcal{O}\left((\delta t)^2\right),
\end{equation}
where $\{\xi, \eta\}$ represent the spatial locations and $\delta t$ is time spacing. In the network implementation, it can be organized as a convolutional kernel $K_t$,
$$
K_t=[-1,0,1] \times \frac{1}{2 \delta t} .
$$
Likewise, we also apply the central difference to calculate the spatial derivatives for internal nodes and use forward/backward differences for boundary nodes. For instance, in this paper, the fourth-order central difference is utilized to approximate the first and second spatial derivatives. The FD kernels for $2 \mathrm{D}$ cases with the shape of $5 \times 5$ are given by
\begin{equation}\label{fixed_operator}
K_{s, 1}=\frac{1}{12(\delta x)}\left[\begin{array}{ccccc}
0 & 0 & 0 & 0 & 0 \\
0 & 0 & 0 & 0 & 0 \\
1 & -8 & 0 & 8 & -1 \\
0 & 0 & 0 & 0 & 0 \\
0 & 0 & 0 & 0 & 0
\end{array}\right], K_{s, 2}=\frac{1}{12(\delta x)^2}\left[\begin{array}{ccccc}
0 & 0 & -1 & 0 & 0 \\
0 & 0 & 16 & 0 & 0 \\
-1 & 16 & -60 & 16 & -1 \\
0 & 0 & 16 & 0 & 0 \\
0 & 0 & -1 & 0 & 0
\end{array}\right],    
\end{equation}
where $\delta x$ denotes the grid size of $\mathrm{HR}$ variables; $K_{s, 1}$ and $K_{s, 2}$ are FD kernels for the first and second derivatives, respectively. In addition, we conduct a parametric study on the selection of FD kernels, including the second-order $(3 \times 3)$, the fourth-order $(5 \times 5)$, and the sixth-order $(7 \times 7)$ central difference strategies.
\subsubsection{Structure-preserved spectral operator}\label{spectral operator}
In Fourier space, the simplification of problem-solving involves converting differential operations and wave number multiplications. In 2D space, the discrete values of the set function $\Phi(x, y)$ in real space, denoted as $\Phi_{ij}$, correspond to $\hat{\Phi}_{mn}$ in Fourier space, with $k_x$ and $k_y$ representing the respective wave numbers. Differential operators are transformed as follows. (1) The first-order differential operator, $\frac{\partial \Phi}{\partial x}$, becomes $i k_x \hat{\Phi}_{mn}$ in the $x$ direction and $i k_y \hat{\Phi}_{mn}$ in the $y$ direction. (2) The second-order differential operator, $\frac{\partial^2 \Phi}{\partial x^2}$, is represented as $-k_x^2 \hat{\Phi}_{mn}$ for the $x$ direction and $-k_y^2 \hat{\Phi}_{mn}$ for the $y$ direction. (3) The Laplacian operator $\nabla^2 \Phi$ transforms into $(-k_x^2 - k_y^2) \hat{\Phi}_{mn}$ in Fourier space.

Moreover, for a 2D flow field, the relationship between the flow function $\psi$ and the velocity fields $(u,v)$ can be expressed by the following partial differential equation, $u = \frac{{\partial \psi}}{{\partial y}}$ and $v = -\frac{{\partial \psi}}{{\partial x}}$, where we can use $k_x$ and $k_y$ to obtain differential results. Moreover, here \textbf{S-Conv} is from FNO~\citep{li2020fourier} (the module named as ``SpectralConv2d\_fast"). This ``SpectralConv2d\_fast" class is a neural network module that performs a 2D spectral convolution by applying an FFT, a learned linear transformation in the Fourier domain, and an IFFT.
\subsection{The Instructions of TSSM}\label{Instructions_TSSM}
Knowing whether a system exhibits \textbf{explicit}, \textbf{implicit}, or \textbf{hybrid} structures is beneficial but not strictly necessary before choosing a structure. Our method provides the flexibility to select an appropriate architecture based on the problem's characteristics, even with partial knowledge. At the same time, complete comprehension of these structural types can guide the choice more precisely. In the setting of our problem, the conservation relation of the process system is of definite form,
$$\frac{\partial \boldsymbol{U}}{\partial t} = -\nabla\cdot (\boldsymbol{J}_{C}+\boldsymbol{J}_{D})+\boldsymbol{q}+\boldsymbol{F}$$
Under such a premise, we can easily select the appropriate structures for different spatio-temporal systems according to the following rules (\textbf{The choice of different paths} and \textbf{Impact of Mismatched Path Selection on Performance}).

\subsubsection{The choice of different paths}
As shown in Tab. \ref{model_types}, we defined three structural types: localized, spectral, and hybrid, each suited to different system characteristics. Localized path is appropriate for systems like the Burgers equation, where diffusion and convection terms are explicit and uncoupled. Spectral path is better for systems with coupled terms, like the vorticity form of the NS equation. Hybrid path suits systems combining these elements, like the general form of the NS equation, where diffusion and convection are uncoupled, but the internal source term is coupled (i.t. the pressure and target velocity fields are coupled to satisfy the Poisson equation). When the interaction between convection and diffusion terms is unknown, the spectral path is a reliable approximation for all these elements, and the hybrid path addresses the source term (Internal/External Source).

\begin{table}[h]
\centering
\renewcommand{\arraystretch}{1}
\caption{The choice of different paths.}
\vspace{2mm}
\label{model_types}
\begin{tabular}{c|c|c|c}
\toprule
\textbf{} & \textbf{Localized} & \textbf{Spectral} & \textbf{Hybrid} \\ 
\midrule
Characteristic & Explicit & Implicit & Explicit+Implicit \\ 
\midrule
Example & $-u\nabla u + \nabla^2 u$ & $-u\nabla w+\nabla^2 w$ & $-u\nabla u + \nabla^2 u - \nabla p$ \\ 
\midrule
Diffusive Flows/Convective Flows & Pre-defined convolution & E-Conv & Pre-defined convolution \\ 
\midrule
Internal Source Term/External Source Term & ResNet block & S-Conv block & ResNet/S-Conv block \\ 
\bottomrule
\end{tabular}
\end{table}

\begin{table}[h]
\centering
\renewcommand{\arraystretch}{1}
\caption{Impact of Mismatched Path Selection on Performance.}
\vspace{2mm}
\label{path_performance}
\begin{tabular}{c|c|c|c|c}
\toprule
\textbf{Datasets} & \textbf{Category} & Localized & Spectral & Hybrid \\ 
\midrule
Burgers2d & Localized & 0.039 & 0.043 & \textbf{0.037} \\ 
\midrule
NS2d ($\nu=1e$-5) & Spectral & 0.061 & \textbf{0.034} & 0.048 \\ 
\midrule
NSM2d & Hybrid & 0.205 & 0.196 & \textbf{0.189} \\ 
\bottomrule
\end{tabular}
\end{table}

\subsubsection{Impact of Mismatched Path Selection on Performance}
Using all three paths, we evaluated the performance across three datasets (Burgers2d for Localized, NS2d for Spectral, and NSM2d for Hybrid). The results are summarized as Tab. \ref{path_performance}. For Burgers2d and NSM2d, with explicit diffusion and convection terms, physical space approximation outperforms spectral space gradient approximation.
For NS2d, featuring implicit diffusion and convection terms, spectral space gradient approximation is superior to physical space approximation. Across all datasets, the Hybrid path surpasses the Localized path in performance. However, this comes at a cost: an increase in model parameters from 14k to 35k, higher computational demands, and marginal performance gains. Consequently, the Localized path is preferred for scenarios without complex couplings, such as Burgers2d.

\begin{table}[h]
\centering
\renewcommand{\arraystretch}{1}
\caption{The difference of five datasets.}\label{dataset_details}
\vspace{2mm}
\begin{tabular}{l|c|c|c|c|c}
\bottomrule
\multirow{2}{*}{Dataset} & \multirow{2}{*}{Category} & \multicolumn{4}{c}{Various Conditions}\\
&& Initial conditions& Boundary conditions& Coefficients& External sources\\
\midrule
Burgers2d&Localized& \checkmark & Periodic & \checkmark & unknown \\
RD2d & Localized  & \checkmark & No-flow Neumann &  & unknown\\
\midrule
NS2d & Spectral  & \checkmark & Periodic   &           & unknown\\
\midrule
Lid2d& Hybrid  &              & Dirichlet, Neumann & \checkmark & \\
NSM2d& Hybrid    & \checkmark & Dirichlet, Neumann & \checkmark & unknown  \\
\bottomrule
\end{tabular}
\end{table}
\section{Datasets}\label{data}
As shown in Tab.\ref{dataset_details}, we employ five datasets spanning diverse domains, such as fluid dynamics and heat conduction. Based on the TSSM scheme employed by PAPM, we categorize the aforementioned five datasets into three types: \textbf{Burgers2d} and \textbf{RD2d} fall under the \textbf{localized} category, \textbf{NS2d} is classified as \textbf{spectral}, while \textbf{Lid2d} and \textbf{NSM2d} are designated as \textbf{hybrid}. The generations of Lid2d and NSM2d are detailed via COMSOL multiphysics in~\ref{comsol}. We are particularly keen to make \textbf{Lid2d} and \textbf{NSM2d} publicly available, anticipating various research endeavors on these datasets by the community.

\subsection{Five datasets}
\textbf{Burgers2d~\citep{huang2023learning}.} The 2D Burgers equation is a fundamental nonlinear partial differential equation. Its formulation is given by:
\begin{equation}
\left\{
\begin{aligned}
\frac{\partial \boldsymbol{u}}{\partial t} & =-\boldsymbol{u} \cdot \nabla \boldsymbol{u}+v \Delta \boldsymbol{u}+\boldsymbol{f}, \\
\left.\boldsymbol{u}\right|_{t=0} & =\boldsymbol{u}_0(x, y)
\end{aligned}
\right.
\end{equation}
where $\boldsymbol{u}=(u(x, y, t), v(x, y, t))$ represents the velocity field, and the spatial domain is $\Omega=[0,2\pi]^2$ with periodic boundary conditions. The viscosity coefficient $v$ varies within the range $v\in[0.001,0.1]$. The forcing term is defined as:
\begin{equation}
f(x, y, \boldsymbol{u})=(\sin (v) \cos (5 x+5 y), \sin (u) \cos (5 x-5 y))^{\top}.
\end{equation}
The initial condition, denoted as $\boldsymbol{u}_0(x, y)$, is drawn from a Gaussian random field characterized by a variance of $25(-\Delta+25 I)^{-3}$. Subsequently, it is linearly normalized to fall within the $[0.1, 1.1]$ range. A total of $N = 250$ samples are generated, each spanning $M = 3200$ time steps with a step size of $\delta t=\frac{0.01}{32}$. A high-resolution traditional numerical solver is employed to generate high-precision numerical solutions. This solver utilizes the $\delta t$ value and operates on a finely discretized $256 \times 256$ grid. The resulting high-precision solutions are stored at intervals of every $32$ time step, resulting in $100$ time slices. Subsequently, these solutions are downsampled to a coarser $64 \times 64$ grid. 

\textbf{RD2d~\citep{takamoto2022pdebench}.} Considering the 2D diffusion-reaction equation, the conservation of the activator $u$ and inhibitor $v$ can be represented as:
\begin{equation}
\left\{
\begin{aligned}
\frac{\partial u }{\partial t} &= -\nabla J_u + R_u, \frac{\partial v }{\partial t} = -\nabla J_v + R_v\\
J_u &= -D_u \nabla u,\quad J_v = -D_v \nabla v
\end{aligned}
\right.
\end{equation}
Where $J_u$ and $J_v$ are the flux terms for the activator and inhibitor, respectively. These represent the diffusive or molecular flows for each component. The reaction functions $R_u$ and $R_v$ for the activator and inhibitor, respectively, are defined by the Fitzhugh-Nagumo (FN) equation, written as $R_u = u - u^3 - k - v$ and $R_v = u - v$, where $k = 5 \times 10^{-3}$ and the diffusion coefficients for the activator and inhibitor are $D_u = 1 \times 10^{-3}$ and $D_v = 5 \times 10^{-3}$, respectively. The initial condition is characterized by a standard normal random noise, with $u(0, x, y) \sim \mathcal{N}(0, 1.0)$ for $x \in (-1, 1)$ and $y \in (-1, 1)$. The boundary conditions are defined as no-flow Neumann boundary conditions. This entails that the partial derivatives satisfy the conditions: $D_u \partial_x u = 0$, $D_v \partial_x v = 0$, $D_u \partial_y u = 0$, and $D_v \partial_y v = 0$, all applicable for the domain $x, y \in (-1, 1)^2$. This dataset~\footnote{This dataset can be downloaded at \href{https://github.com/pdebench/PDEBench}{https://github.com/pdebench/PDEBench}} is transformed into a coarser grid with dimensions of $64 \times 64$ while keeping the time step consistently constant.

\textbf{NS2d~\citep{li2020fourier}.} We refer to \textbf{FNO} as the source for our exploration of the two-dimensional incompressible Navier-Stokes equation in vorticity form. This equation is defined on the unit torus and is outlined as follows:
\begin{equation}
\left\{
\begin{aligned}
\partial_t w(x, t) + u(x, t) \cdot \nabla w(x, t) &= \nu \Delta w(x, t) + f,  \\
\nabla \cdot u(x, t) &= 0,  \\
w(x, 0) &= w_0(x),
\end{aligned}
\right.
\end{equation}
where $x \in (0,1)^2$, $t \in (0, T]$, and $u$ represents the velocity field, $w = \nabla \times u$ denotes the vorticity, $w_0$ stands for the initial vorticity distribution, $\nu \in \mathbb{R}_{+}$ signifies the viscosity coefficient and $f$ denotes the forcing function. In this work, the viscosity coefficient is set to $\nu = 1 \times 10^{-3}, 1 \times 10^{-4}, 1 \times 10^{-5}$. It's worth noting that, for the purpose of maintaining a consistent evaluation framework, the resolution is standardized at $64 \times 64$ for both training and testing phases, given that the baseline methods are not inherently resolution-invariant.

\textbf{Lid2d.} A constant velocity across the top of the cavity creates a circulating flow inside. To simulate this, a constant velocity boundary condition is applied to the lid while the other three walls obey the no-slip condition. Different Reynolds numbers yield different results, so in this article, $Re \in[100, 1500]$ are applied. At high Reynolds numbers, secondary circulation zones are expected to form in the corners of the cavity. The system of differential equations (N-S equations) consists of two equations for the velocity components $\boldsymbol{u}=(u(x, y, t), v(x, y, t))$, and one equation for pressure $(p(x, y, t))$:
\begin{equation}
\left\{
\begin{aligned}
\frac{\partial \boldsymbol{u}}{\partial t} & =-\boldsymbol{u} \cdot \nabla \boldsymbol{u}+\frac{1}{R e} \Delta \boldsymbol{u}-\nabla p, \\
\nabla \cdot \boldsymbol{u} & =0
\end{aligned}
\right.
\end{equation}
where $(x,y) \in (0,1)^2$. The initial condition is $(u, v, p)=\boldsymbol{0}$ everywhere. And the boundary conditions are: $u=1$ at $y=1$~(the lid), $(u,v)=\boldsymbol{0}$ on the other boundaries, $\partial p/\partial y=0$ at y=0$, $p=0 at $y=1$, and $\partial p/\partial x=0$ at $x=0,1$. The data generation for the Lid2d is processed by COMSOL Multiphysics®, and a total of $N = 200$ samples are generated, each spanning $M = 1000$ time steps with a step size of $\delta t=\frac{0.1}{10}$. Every 10 steps, we save the data, resulting in $100$ time slices. This solver utilizes the value of $\delta t$ and operates on a finely discretized $128 \times 128$ grid. Subsequently, these solutions are downsampled to a coarser $64 \times 64$ grid. 

\textbf{NSM2d.} Consider the Navier-Stokes equations with an additional magnetic field:
\begin{equation}
\left\{
\begin{aligned}
\frac{\partial \boldsymbol{u}}{\partial t}+\boldsymbol{u} \cdot \nabla \boldsymbol{u} & =-\nabla p+\nu \nabla^2 \boldsymbol{u}+\boldsymbol{F}, \quad t \in[0, T], \\
\nabla \cdot \boldsymbol{u} & =0,
\end{aligned}    
\right.
\end{equation}
where $(x,y)\in [0,4]\times[0,1]$, $\boldsymbol{u}=[u(x, y, t), v(x, y, t)] \in \mathbb{R}^2$ is the velocity vector, $p(x, y, t) \in \mathbb{R}$ is the pressure, $\nu=1 / R e$ represents the kinematic viscosity (with $R e$ as the Reynolds number), and $\boldsymbol{F}=\left[F_x, F_y\right]$ is an external source term induced by the magnetic field. The components of $\boldsymbol{F}$ are defined as follows:
\begin{equation}
\left\{
\begin{aligned}
F_x & =m H \frac{\partial H}{\partial x}, \quad F_y=m H \frac{\partial H}{\partial y} \\
H(x, y) & =\exp \left[-8\left((x-L / 2)^2+(y-W / 2)^2\right)\right]
\end{aligned}
\right.
\end{equation}
where $L=4$, $W=1$, $m=0.16$ is the magnetization, and $H$ is a time-invariant magnetic intensity. The simulation is conducted on a $2 \mathrm{D}$ rectangular domain $\{x, y\} \in[0,4] \times[0,1]$ with the following boundary conditions: the inflow boundary $(x=0)$ is prescribed with a velocity distribution $\boldsymbol{u}(0, y, t)$, where $y_0$ represents the vertical position of the inlet jet center:
\begin{equation}
\boldsymbol{u}(0, y, t)=\left[\begin{array}{l}
u(0, y, t) \\
v(0, y, t)
\end{array}\right]=\left[\begin{array}{c}
\exp \left(-50\left(y-y_0\right)^2\right) \\
\sin (t) \cdot \exp \left(-50\left(y-y_0\right)^2\right)
\end{array}\right]
\end{equation}
The outflow boundary $(x=4)$ is set with a reference pressure $p(4, y, t)=0$. The no-slip boundary condition is applied at the top and bottom walls ($y=0, 1$). The Reynolds number is dimensionless and ranges from $100$ to $1500$. The inlet jet position $y_0$ is varied within the domain $0.4 \leq y_0 \leq 0.6$. The data generation for the NSM2d is processed by COMSOL Multiphysics®, and a total of $N = 200$ samples are generated, each spanning $M =1000$ time steps with a step size of $\delta t=\frac{0.2}{10}$. Every 10 steps, we save the data, resulting in $100$ time slices. This solver utilizes the $\delta t$ value and operates on a finely discretized $256 \times 64$ grid. Subsequently, these solutions are downsampled to a coarser $128 \times 32$ grid. 

\subsection{Detailed Data Generation Process}\label{comsol}
Our research employed COMSOL multiphysics software~\footnote{\href{https://www.comsol.com/}{https://www.comsol.com/}} for fluid dynamics simulation in a lid-driven cavity and a magnetic stirring scenario. The main text outlines the simulation parameters, utilizing grids of $128 \times 128$ and $256 \times 64$ for each case, respectively. The Time-Dependent Module, with specific time steps, was used for execution. The simulations required substantial computational resources, solving for 49,152 and 16,130 internal degrees of freedom (DOFs) in each scenario. To generate a comprehensive dataset, we varied simulation parameters, running $200$ simulations for each scenario with different Reynolds numbers and, in the magnetic stirring case, the $y_0$ value. This data was stored in h5 format.

The computational intensity was significant: a single run in the lid-driven scenario took 91 seconds on average, while the magnetic stirring case took 226 seconds. The total computation time was approximately $10^5$ seconds for all 200 cases, highlighting the time-consuming nature of such simulations. The intricacy of multi-physics coupling and the extensive computational demand in these simulations point towards the necessity of more efficient methods. This situation underscores the potential of neural networks in accelerating simulation processes. By leveraging neural networks, we aim to reduce the computational time significantly, addressing the inherent slowness of detailed simulations like those in our study. This approach could transform the feasibility and scalability of complex simulations in various scientific and engineering domains.

\section{Details for experimental setup and models' hyper-parameters}\label{all_details}

\subsection{Experimental setup}
Based on the experimental setting of time extrapolation, we further conducted experiments in the following two parts: coefficient interpolation (referred to as \textbf{C Int.}) and coefficient extrapolation (referred to as \textbf{C Ext.}). 
For \textbf{C Int.}, the data is uniformly shuffled and then split into training, validation, and testing datasets in a $[7:1:2]$ ratio. 
However, in the case of \textbf{C Ext.}, the data splitter is determined based on the order of coefficients, with equal proportions $[7:1:2]$. For example, the viscosity coefficients are divided from largest to smallest, and the coefficients with the lowest viscosity, representing the most challenging tasks, are selected as the test set. 

One crucial point to consider is that to maintain consistency with the data-driven approach~\cite{shi2015convolutional, stachenfeld2021learned, li2020fourier, gupta2022towards, raonic2023convolutional}, we must replace the initial time $t=0$ with the initial step size $t_0$.
We consistently set the initial time step size for all datasets across various tasks as $t_0 = 5$. In the test set, $T_{end} = 100$, except for the NS2d. Specifically, for NS2d with viscosity values of $\nu=1\mathrm{e}{-3}$ and $1\mathrm{e}{-4}$, $T_{end} = 50$, while for $\nu = 1\mathrm{e}{-5}$, $T_{end} = 20$. The trajectory length in the training set that can be used as label data is given by $T_{end}/2 - t_0$. 

We train all models with AdamW~\citep{loshchilov2017decoupled} optimizer with the exponential decaying strategy, and epochs are set as $500$. The causality parameter $\alpha_1=0.1$ and $\alpha_0=0.001$. The initial learning rate is $1e$-3, and the ReduceLRonPlateau schedule is utilized with a patience of $20$ epochs and a decay factor of $0.8$. For a fair comparison, the batch size is identical across all methods for the same task, and all experiments are run on $1\sim3$ NVIDIA Tesla P100 GPUs.

To account for potential variability due to the partitioning process, each experiment is performed three times, and the final result is derived as the average of these three independent runs. Except for the predefined parameters, the parameters of all models are initialized by Xavier~\citep{glorot2010understanding}, setting the scaling ratio $c=0.02$.

\subsection{Hyper-parameters}\label{details}
\textbf{PAPM.}  In this work, PAPM designed three different temporal-spatial modeling methods according to the characteristics of five different data sets. 
\begin{itemize}[itemsep=1pt,topsep=0pt,leftmargin=*]
    \item \textbf{Localized Operator.} Burgers2d uses the predefined fixed convolution kernel as the convolution kernel parameter of diffusive and convective flows, while a $4$-layer convolution layer characterizes the source term. Its channel is set to $16$, and the GELU activation function is used. In RD2d, trainable convolutional kernels are used as the convolution kernel parameter of diffusive flows, and the kernel is set to $5$. For the source term, like burgers2d, a four-layer convolutional layer with channel $16$ and GELU is used to characterize the source term.
    \item \textbf{Spectral Operator.} After FFT, $k_x$, $k_y$ and $\hat{w}$ are input into a $1\times1$ conv for dot product, which is used to solve the partial derivatives of vorticity $w$ and velocity field $u$, and then to physical space through IFFT. Simple operations such as multiplication and addition are performed according to conservation relations. As for the source term is characterized by a layer of S-Conv (\textit{i.e.}, spectralConv2d\_fast) with (width$=12$, modes1$=12$, modes2$=12$).
    \item \textbf{Hybrid Operator.} According to the velocity part of the conservation equation, we set the kernel as five using trainable convolutional kernels as the convolution kernel parameters of diffusive and convective flows. We use a three-layer convolutional layer with channel $16$ and GELU to represent the source term. Then, the intermediate results of the velocity field are fed into an S-Conv (width$=8$, modes1$=8$, modes2$=8$) to map the complete velocity field and pressure field.
\end{itemize}

\textbf{ConvLSTM~\citep{shi2015convolutional}}. Specializing in spatial-temporal prediction, ConvLSTM blends LSTM's temporal cells with CNN spatial extraction. The setup consists of three distinct blocks: an encoding block employing a $5 \times 5$ convolution kernel with channel $32$, an LSTM cell-based forecasting block, and a decoding block featuring $2$ CNN layers with $5 \times 5$ kernels and $2$ Res blocks. Notably, multi-step predictions are achieved through state concatenation within the forecasting block. Despite its strengths, its performance in complex process systems can be limited due to potential error accumulations.

\textbf{Dil-ResNet~\citep{stachenfeld2021learned}}. This model combines the encode-process-decode paradigm with the dilated convolutional network. The processor consists of $N=4$ residual blocks connected in series, and each is made of dilated CNN stacks with residual connections. One stack consists of $7$ dilated CNN layers with dilation rates of $(1, 2, 4, 8, 4, 2, 1)$, where a dilated rate of $N$ indicates that each pixel is convolved with multiples of $N$ pixels away. Each CNN layer in the processor is followed by ReLU activation. The key part of this network is a residual connection, which helps avoid vanishing gradients, and dilations allow long-range communication while preserving local structure. We found difficulties running this model on complex datasets due to computing and memory constraints. 

\textbf{time-FNO2D~\citep{li2020fourier}}. This model applies matrix multiplications in the spectral space with learnable complex weights for each component and linear updates and combines embedding in the spatial domain.  The model consists of $2$ MLP layers for encoding and decoding and $4$ Fourier operation blocks (width$=12$, modes1$=12$, modes2$=12$). Each block contains Fourier and CNN layers, followed by GELU activation. Optionally, low-pass filtering truncates high-frequency modes along each dimension in the Fourier-transformed grid. 

\textbf{MIONet~\citep{jin2022mionet}}. In the original paper, the Depth of MIONet is set to 2, the width is 200, and the number of parameters is 161K. Build MIONet with DeepXDE~\citep{lu2021deepxde}~\footnote{\href{https://github.com/lululxvi/deepxde}{https://github.com/lululxvi/deepxde}}. In this work, we have greatly adjusted the number of parameters; the number of parameters is about 20k, and the width is set to 20. Consistent with FNO, MLP is also introduced to construct the project layer for data, and then projection is also carried out in output. Other contents are consistent with the original text.

\textbf{U-FNet~\citep{gupta2022towards}}. This model improves U-Net architectures, replacing lower blocks both in the downsampling and in the upsampling path of U-Net architectures by Fourier blocks, where each block consists of 2 FNO layers and residual connections. Other contents are consistent with the original text. In this work, we have adjusted $n\_input\_scalar\_components$ and $n\_output\_scalar\_components$ to the number of channels of the physical field in our datasets, and both $time\_history$ and $time\_future$ are set to 5 for better fitting.

\textbf{CNO~\citep{raonic2023convolutional}}.This model proposes a sequence of layers with the convolutional neural operator, mapping between bandlimited functions based on U-Net architectures. The convolutional neural operator consists of 4 different blocks, i.e., the downsampling block, the upsampling block, the invariant block, and the ResNet block. In the original paper, the width and height of spatial size for the mesh grid should be identical. In this work, we relaxed this restriction in activation function $filtered\_lrelu$ to fit on non-square grids like the NSM2d dataset. Other contents are consistent with the original text.

\textbf{PeRCNN~\citep{rao2023encoding}}. The network consists of two components: a fully convolutional decoder as an initial state generator and a novel recurrent block named $\prod$-block for recursively updating the state variables. Since our experiments' available measurement size is full, we have omitted this decoder. In the recurrent $\prod$-block, the state first goes through multiple parallel $1 \times 1$ Conv layers with stride $1$ and output channel $32$. The feature maps produced by these layers are then fused via the elementwise product operation. Then, the multi-channel goes through a conv layer with a filter size of $1$ to obtain the output of the desired number of channels. We found this method unstable when approaching nonlinear complex terms and prone to NaN values during training.

\textbf{PPNN~\citep{liu2024multi}}. This model combines known partial nonlinear functions with a trainable neural network, which is named ConvResNet. The only difference between these two models is that a trainable portion of PPNN has an extra input variable $\mathcal{F}$, provided by the PDE-preserving portion of PPNN. The state first goes through the decoder, which is made of four ConvResNet blocks, and each of them consists of a $7\times7$ kernel with $96$ channels and a zero padding of $3$. The following decoder includes a pixel shuffle with an upscale factor equal to $4$ and a convolution layer with a $5\times5$ kernel. Due to the physics-aware design, this model shows lower relative error in the extrapolation range.

\section{Additional experimental results}\label{additional_results}
In this section, some additional experimental results are shown, which are data visualization (\ref{visual}), training and inference time cost-specific details (\ref{cost}), and detailed ablation studies (\ref{supp_ablation}).

\subsection{Visualization}\label{visual}
\begin{figure}[h]
\centering
\includegraphics[width=\textwidth]{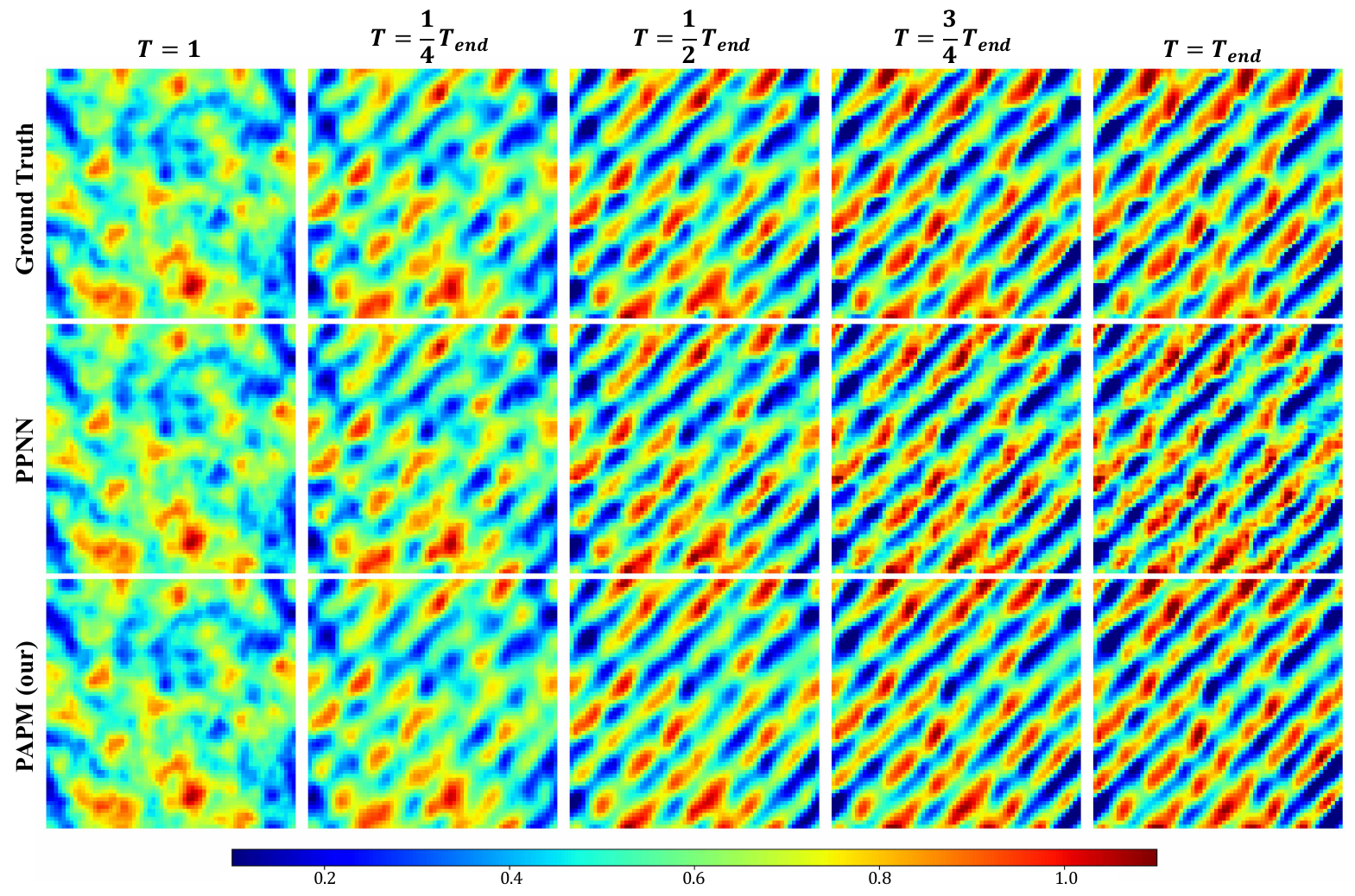}
\vspace{-4mm}
\caption{Predicted flow velocity ($\| \boldsymbol{u} \|_2$) snapshots by PPNN, and PAPM (Ours) vs. Ground Truth (GT) on Burgers2d dataset in T Ext. task.}\label{fig_bur}
\vspace{-2mm}
\end{figure}

\begin{figure}[h]
\centering
\includegraphics[width=\textwidth]{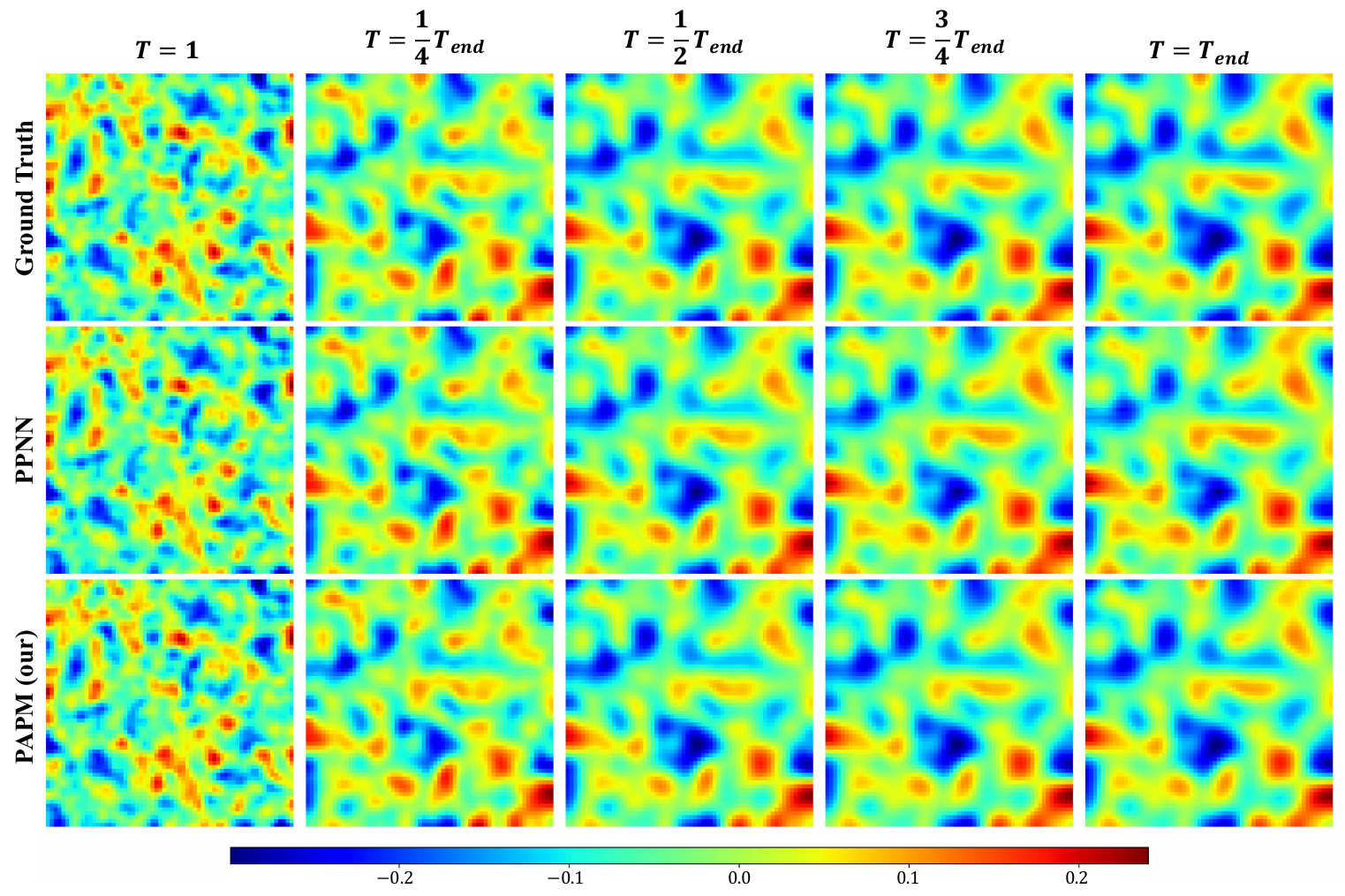}
\vspace{-2mm}
\caption{Predicted flow velocity ($\| \boldsymbol{u}\|_2$) snapshots by PPNN, and PAPM (Ours) vs. Ground Truth (GT) on RD2d dataset in T Ext. task.}\label{fig_rd}
\vspace{-4mm}
\end{figure}

\begin{figure*}[h]
\centering
\includegraphics[width=\textwidth]{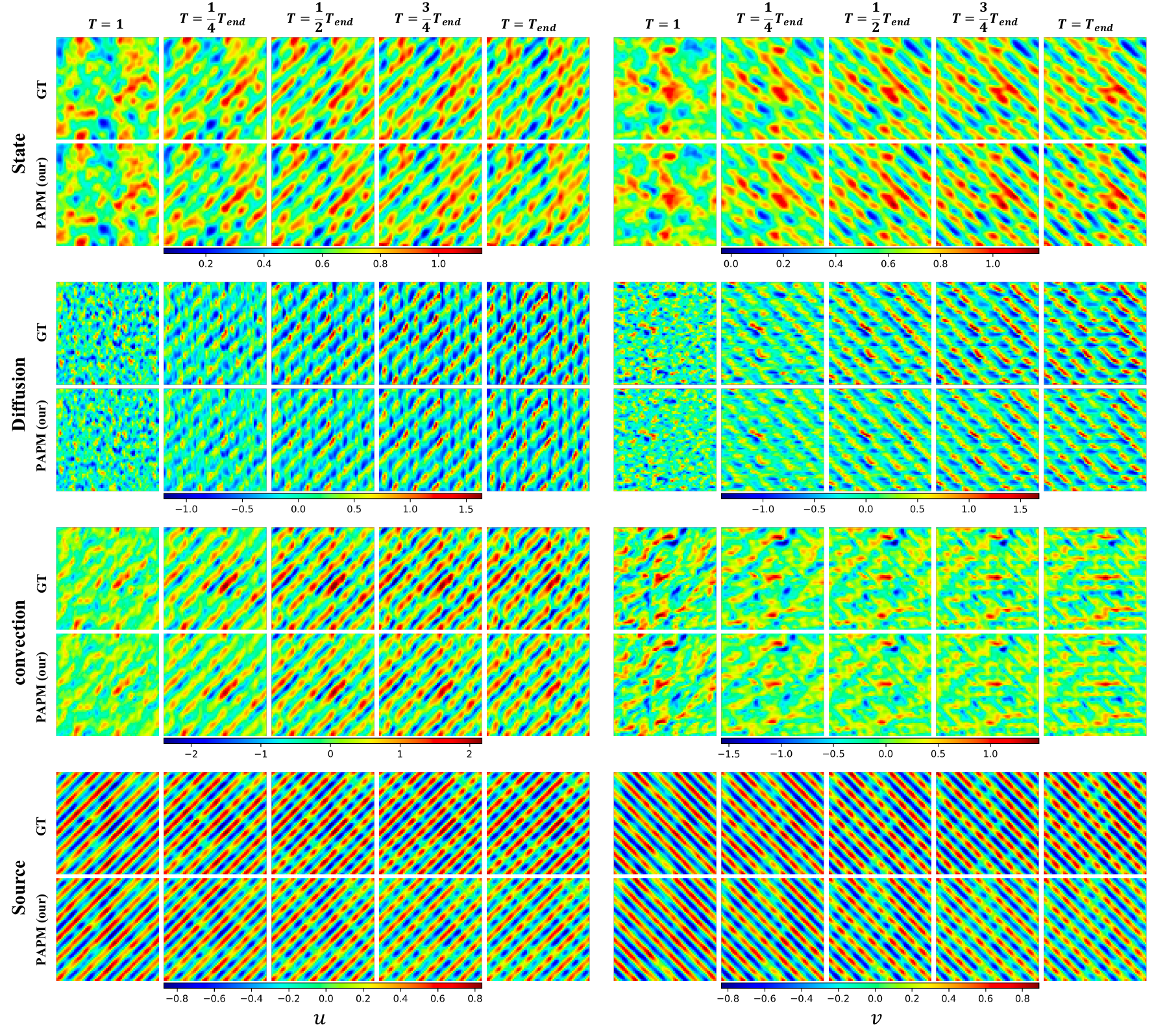}
\vspace{-4mm}
\caption{Different terms on the Burgers2d dataset in T Ext. task.}\label{bur_ab}
\vspace{-2mm}
\end{figure*}

\begin{figure*}[h]
\centering
\includegraphics[width=\textwidth]{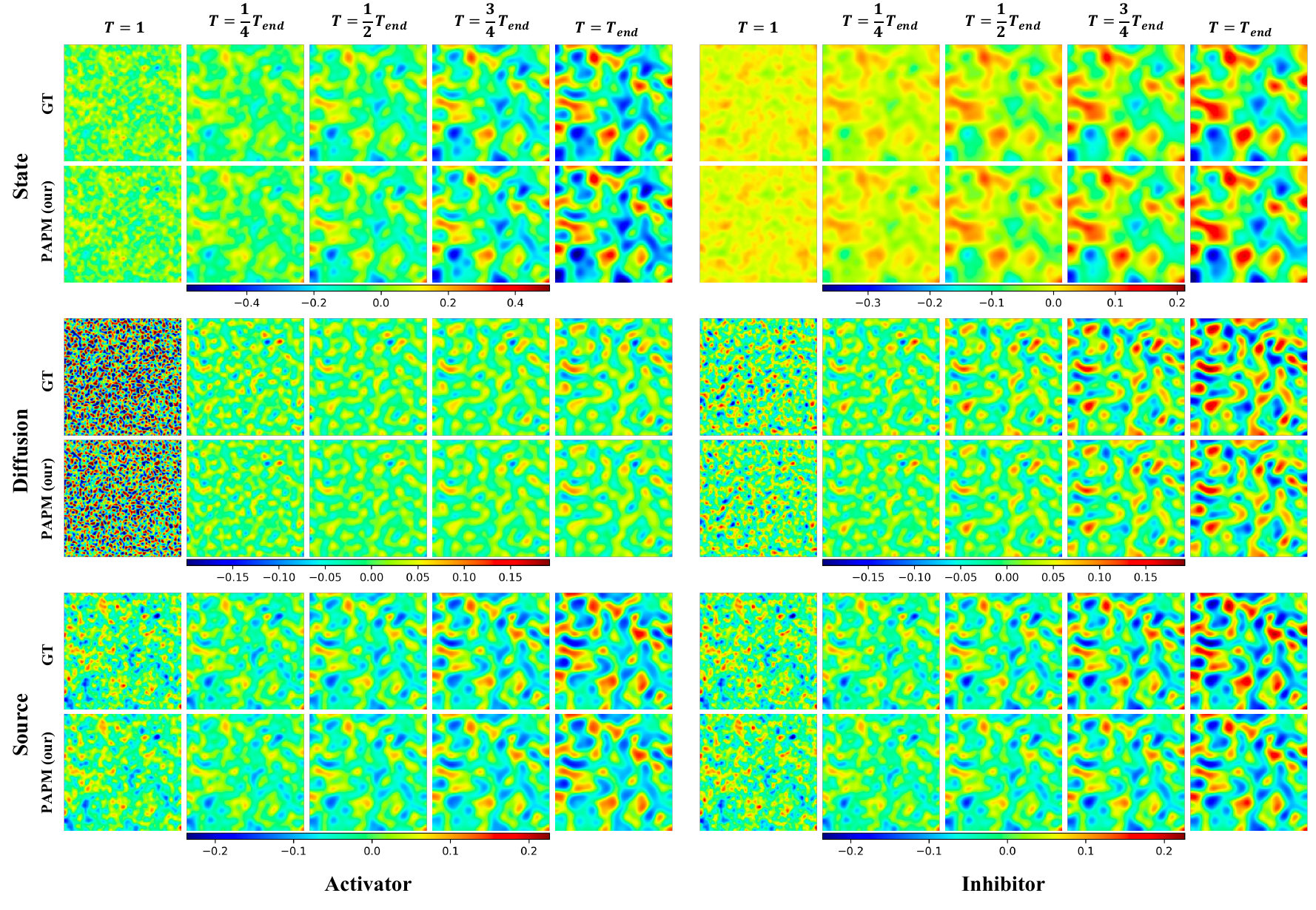}
\vspace{-4mm}
\caption{Different terms on the RD2d dataset in T Ext. task.}\label{rd_ab}
\vspace{-2mm}
\end{figure*}

Fig.~\ref{fig_bur} and Fig.~\ref{fig_rd} showcase the results across five extrapolation time slices on Burgers2d and RD2d datasets. Both datasets clearly show that the physics-aware methods, PPNN and PAPM (our), can predict the dynamics of these two complex systems well. However, in the second half of the extrapolation ($T\geq \frac{1}{2}T_{end}$), It can be seen that our method PAPM is better than PPNN in local detail reconstruction. It is worth mentioning that our method has only 1\% of the number of parameters and FLOPs of PPNN. However, the effect is better than PPNN, which further affirms the superiority of our structured design and specific spatio-temporal modeling method. Fig.~\ref{bur_ab} and Fig.~\ref{rd_ab} show the visual effects between different terms of PAPM and the numerical results, which can further prove that PAPM can learn the equation's convection/diffusion/source parts.

\subsection{Training and inference time cost}\label{cost}
Dataset generation for our work is notably resource-intensive, with inference costs ranging from $10^2\sim10^4$ s for public datasets and up to $10^3$ s for those we generated using COMSOL Multiphysics®. As detailed in Tab.~\ref{per_time}, in stark contrast, both baselines and PAPM register inference times between $0.1\sim10$ s, achieving an improvement of 3 to 5 orders of magnitude. Notably, PAPM's time cost rivals or even surpasses baselines across different datasets. Unlike traditional numerical algorithms, this indicates that introducing rigorous physical mechanisms doesn't necessarily bloat time costs. PAPM's efficiency remains competitive with other data-driven methods.
\begin{table*}[h]%
\centering
\renewcommand{\arraystretch}{1}
\caption{Training and inference time cost (epoch/second) of different baselines.}
\label{per_time}
\vspace{2mm}
\small{
\begin{tabular}{l|cc|cc|cc|cc|cc}
    \toprule
    \multirow{2}{*}{\textbf{Config}}  
    & \multicolumn{2}{c|}{\textbf{Burgers2d}} 
    & \multicolumn{2}{c|}{\textbf{RD2d}}
    & \multicolumn{2}{c|}{\textbf{NS2d}}
    & \multicolumn{2}{c|}{\textbf{Lid2d}}
    & \multicolumn{2}{c}{\textbf{NSM2d}}\\ 
    &Train&Infer
    &Train&Infer
    &Train&Infer
    &Train&Infer
    &Train&Infer\\
    \midrule
    ConvLSTM    &5.41&1.12&21.41&3.94&7.05&0.86&\textbf{8.68}&3.22&4.39&0.92\\ 
    Dil-ResNet  &6.64&1.73&27.06&4.06&9.96&1.19&10.90&3.66&6.34&1.03\\ 
    time-FNO2D  &4.87&1.46&8.94&1.95&5.16&0.79&10.41&2.11&\textbf{3.35}&0.69\\ 
    MIONet      &5.69&1.58&8.69&2.03&5.05&0.89&10.54&3.02&\underline{4.03}&0.76\\
    {U-FNet} 
    &{\underline{3.64}}&{\textbf{0.52}}
    &{14.56}&{1.96}
    &{6.67}&{\underline{0.51}}
    &{10.42}&{\underline{1.14}}
    &{6.96}&{0.82}\\
    {CNO} 
    &{4.02}&{\underline{0.60}}
    &{15.72}&{2.28}
    &{4.92}&{\textbf{0.44}}
    &{11.08}&{\textbf{1.12}}
    &{5.90}&{\underline{0.68}}\\
    PeRCNN      
    &5.02&1.72&\textbf{5.73}&\underline{1.47}
    &6.53&0.84&17.44&4.08&4.24&0.82\\
    PPNN        
    &5.07&0.96&8.88&\textbf{1.19}&\underline{4.87}
    &0.91&15.58&3.44&8.08&\textbf{0.64}\\ 
    \midrule
    PAPM        
    &\textbf{3.44}&0.93
    &\underline{8.62}&2.07
    &\textbf{3.70}&1.27
    &\underline{8.91}&2.94
    &5.13&0.88\\ 
    \bottomrule
\end{tabular}
}
\end{table*}
\subsection{Supplemental Ablation studies}\label{supp_ablation}

\subsubsection{The comparison between fixed and trainable convolutional kernels} In PAPM, the pre-defined convolutional kernels contain the fixed and trainable ones. For the fixed ones, we can directly use the difference scheme as the parameter of the convolution kernel without participating in the subsequent training optimization. For the trainable one, the trainable parameters in kernels are constrained as the triangular and symmetric matrices, which correspond to unidirectional convection and directionless diffusion. The use of trainable ones is due to spatio-temporal discretization, where the original difference scheme is insufficient in characterizing the corresponding gradient information. Thus, we make the kernel in the difference scheme in Diffusive Flows (DF) and Convective Flows (CF) learnable for better capturing spatial gradients. 

\begin{table}[h]
\centering
\caption{Performance comparison for different configurations.}
\vspace{2mm}
\renewcommand{\arraystretch}{1}
\begin{tabular}{c|c|c}
\toprule
\textbf{config} & \textbf{Burgers2d} & \textbf{RD2d} \\ \midrule
fixed & 0.082 & 0.049 \\ \midrule
trainable & \textbf{0.039} & \textbf{0.018} \\ 
\bottomrule
\end{tabular}
\label{kernel_comparison}
\end{table}

Taking the Burgers2d and RD2d datasets in the C Int. setting as examples to show the different performance for fixed and trainable convolutional kernels. Tab. \ref{kernel_comparison} indicates that when the kernel becomes a learnable term, the model's performance greatly improves, where the metric is the mean $L_2$ relative error (Eq. 4), lower values indicate better results, and \textbf{Bold} indicates the best performance.

\subsubsection{Ablation Studies of Different Terms} 
Here, we present a series of ablation experiments conducted on the Burgers2d, RD2d, and NSM2d datasets within the C Int. setting. Each experiment investigates the impact of excluding specific terms: \textbf{no\_DF}, which omits diffusion; \textbf{no\_CF}, which excludes convection; \textbf{no\_Phy}, which removes both diffusion and convection; \textbf{no\_IST}, which excludes internal sources; \textbf{no\_EST}, which excludes external sources; and \textbf{no\_BCs}, which eliminates the explicit embedding of boundary conditions.

We can get the following insights as shown in Tab. \ref{sup_abl_three_datasets},  where the metric is the mean $L_2$ relative error (Eq. 4), lower values indicate better results, and \textbf{Bold} indicates the best performance. 
\textbf{First}, for Burgers2d and NSM2d, the no\_DF configuration demonstrated the importance of integrating the viscosity coefficient with the diffusion term, with its absence leading to significant errors. 
\textbf{Second}, for Burgers2d and RD2d, internal sources primarily drive the state update process. In NSM2d, both the internal source (gradient of pressure, $\nabla p$) and the external source (time-invariant magnetic intensity) play crucial roles. 
\textbf{Third}, the necessity of adhering to physical laws in boundary conditions was highlighted in the no\_BCs, notably reducing errors.
\begin{table}[h]
\centering
\caption{Performance metrics across different datasets with various modifications.}
\vspace{2mm}
\begin{tabular}{c|c|c|c|c|c|c|c}
\toprule
\textbf{Datasets} & \boldmath{$\epsilon$} & \textbf{no\_DF} & \textbf{no\_CF} & \textbf{no\_Phy} & \textbf{no\_IST} & \textbf{no\_EST} & \textbf{no\_BCs} \\ \midrule
Burgers2d & \textbf{0.039} & 0.067 & 0.062 & 0.149 & 0.174 & - & 0.068 \\ 
\midrule
RD2d & \textbf{0.018} & 0.102 & - & 0.102 & 0.281 & - & 0.083 \\
\midrule
NSM2d & \textbf{0.189} & 0.273 & 0.212 & 0.299 & 0.392 & 0.311 & 0.201 \\ 
\bottomrule
\end{tabular}
\label{sup_abl_three_datasets}
\end{table}

\end{document}

%% file: tssm_table.tex
\begin{table*}[t]
\vspace{-2mm}
\centering
\renewcommand{\arraystretch}{1}
\setlength{\tabcolsep}{8pt}
\caption{Temporal-Spatial Stepping Module (TSSM).}
\small{
\begin{tabular}{cc|ccc}
\toprule
\multicolumn{2}{c|}{\textbf{Category}} &
  \multicolumn{1}{c|}{\begin{tabular}[c]{@{}c@{}}\textbf{Localized}\\ Fig.\ref{TSSM} (Left and Mid), Alg.~\ref{alg1_localized}\end{tabular}} &
  \multicolumn{1}{c|}{\begin{tabular}[c]{@{}c@{}}\textbf{Spectral}\\ Fig.\ref{TSSM} (Right), Alg.~\ref{alg2_spectral}\end{tabular}} &
  \textbf{Hybrid} \\ 
\midrule
\multicolumn{2}{c|}{Characteristic} &
  \multicolumn{1}{c|}{Explicit} &
  \multicolumn{1}{c|}{Implicit} &
  Explicit+Implicit \\ 
\midrule
\multicolumn{2}{c|}{Example} &
  \multicolumn{1}{c|}{$-u\nabla u + \nabla^2 u$} &
  \multicolumn{1}{c|}{$-u\nabla w+\nabla^2 w$} &
  $-u\nabla u + \nabla^2 u - \nabla p$ \\ 
\midrule
\multicolumn{1}{c|}{Temporal} &
  ODE solver &
  \multicolumn{3}{c}{Neural ODE} \\ 
\midrule
\multicolumn{1}{c|}{\multirow{3}{*}{Spatial}} &
    DF &
  \multicolumn{1}{c|}{\multirow{2}{*}{\begin{tabular}[c]{@{}c@{}}Pre-defined convolution\end{tabular}}} &
  \multicolumn{1}{c|}{\multirow{2}{*}{E-Conv}} &
  \multirow{2}{*}{\begin{tabular}[c]{@{}c@{}}Pre-defined convolution\end{tabular}} \\ 
\multicolumn{1}{c|}{} &
  CF &
  \multicolumn{1}{c|}{} &
  \multicolumn{1}{c|}{} &
   \\ 
\cmidrule{2-5} 
\multicolumn{1}{c|}{} &
  IST/EST &
  \multicolumn{1}{c|}{ResNet block} &
  \multicolumn{1}{c|}{S-Conv block} &
  ResNet/S-Conv block \\ 
\bottomrule
\end{tabular}
}
\label{tssm_tab}
\end{table*}

%% file: table_1_all.tex
\begin{table*}[t]%
\centering
\begin{minipage}[t]{0.6\textwidth}
\centering
\renewcommand{\arraystretch}{1}
\setlength{\tabcolsep}{4pt}
\caption{$\epsilon$ (Eq. 4) across different datasets in time extrapolation task.}\label{main_results}
\vspace{2mm}
\scriptsize{
\begin{tabular}{l|c|c|c|c|c|c|c|c|c}
\toprule
\multirow{2}{*}{\textbf{Config}} 
& \multicolumn{2}{c|}{\textbf{Burgers2d}} 
& \multicolumn{1}{c|}{\textbf{RD2d}} 
& \multicolumn{3}{c|}{\textbf{NS2d}} 
& \multicolumn{1}{c|}{\textbf{Lid2d}} 
& \multicolumn{2}{c}{\textbf{NSM2d}} \\
&\text{C Int.} & \text{C Ext.}
&\text{C Int.} 
& \text{$\nu$=$1e$-3} & \text{$\nu$=$1e$-4 } & \text{$\nu$=$1e$-5}
&\text{C Ext.} 
&\text{C Int.}&\text{C Ext.}\\
\midrule                  
ConvLSTM   &0.314&0.551&0.815&0.781&0.877&0.788&1.323&0.910&1.102\\
Dil-ResNet &0.071&0.136&\underline{0.021}&0.152&0.511&0.199&0.261&0.288&0.314\\ 
time-FNO2D &0.173&0.233&0.333&\underline{0.118}&\underline{0.100}&\underline{0.033}&0.265&0.341&0.443\\ 
MIONet&0.181&0.212&0.247&0.139&0.114&0.051&0.221&0.268&0.440\\ 
{U-FNet}& {0.109}& {0.433}& {0.239}& {0.191}& {0.190}& {0.256}& {0.192}& {0.257}& {0.457} \\
{CNO}& {0.112}& {\underline{0.126}}& {0.258}& {0.125}& {0.148}& {\textbf{0.030}}& {0.218}& {\underline{0.197}}& {0.355} \\
PeRCNN &0.212&0.282&0.773&0.571&0.591&0.275&0.534&0.493&0.493\\ 
PPNN&\underline{0.047}&0.132&0.030&0.365&0.357&0.046&\underline{0.163}&0.206&\underline{0.264}\\ 
\midrule
\textbf{PAPM (Our)}       &\textbf{0.039}&\textbf{0.101}&\textbf{0.018}&\textbf{0.110}&\textbf{0.097}&0.034&\textbf{0.160}&\textbf{0.189}&\textbf{0.245}\\ 
\bottomrule
\end{tabular}}
\end{minipage}
\hfill 
\begin{minipage}[t]{0.39\textwidth}
\centering
\renewcommand{\arraystretch}{1}
\setlength{\tabcolsep}{3pt}
\caption{Comparison of parameters and FLOPs.}\label{parameters}
\vspace{2mm}
\scriptsize{
\begin{tabular}{l|c|c|c|c}
\toprule
\multirow{2}{*}{\textbf{Config}} 
    & \multicolumn{3}{c|}{$N_P$}&\multirow{2}{*}{FLOPs/M}\\
&Localized/M & Spectra/M & Hybrid/M &\\
\midrule
ConvLSTM   &0.175& 0.139& 0.211& 32.75\\ 
Dil-ResNet &0.150& 0.148& 0.152& 62.40\\ 
time-FNO2D &0.464& 0.463& 0.464& 6.88\\ 
MIONet     &0.261& 0.261& 0.261& 10.01\\ 
U-FNet &9.853&9.851&9.854& 559.89\\
CNO &2.606&2.600&2.612&835.37\\
PeRCNN&\textbf{0.001}&\textbf{0.001}&\textbf{0.001}& \underline{3.44}\\ 
PPNN&1.201&1.190&1.213&348.56\\ 
\midrule
\textbf{PAPM}&\underline{0.014}&\underline{0.034}&\underline{0.035}&\textbf{1.23}\\
\bottomrule
\end{tabular}}
\end{minipage}
\end{table*}

%% file: notations.tex
\begin{table*}[t]
\centering
\renewcommand{\arraystretch}{1}
\setlength{\tabcolsep}{4pt}
\caption{Table of notations.}
\vspace{2mm}
\small{
\begin{tabular}{lc}
\toprule
\textbf{Notation}         & \textbf{Meaning}        \\ 
\midrule
Process model&\\ 
\midrule
$\boldsymbol{U}$&the physical quantity, also as the system's state\\
$\boldsymbol{J}_{D}$ & the diffusion flows in the conservation relations\\
$\boldsymbol{J}_{C}$& the convection flows in the conservation relations\\
$\boldsymbol{q}$& the internal source in the conservation relations\\
$\boldsymbol{F}$& the external source in the conservation relations\\
$\boldsymbol{v}$ & the velocity of the physical quantity being transmitted \\
$\boldsymbol{D}$ & the coefficients\\
$\boldsymbol{\lambda}$ & the coefficients, such as viscosity \\
$\boldsymbol{X}_{F}$ & a vector of external sources, such as voltages \\
$\boldsymbol{U}^0$&the initial condition\\
ICs&Initial conditions\\
BCs&Boundary conditions\\
\midrule
Problem formulation&\\
\midrule
$\boldsymbol{U}_k^{t} = \{\boldsymbol{u}_{k, i}^{t}\}_{1\leq i \leq m}$ &  a vector, which consists of $m \in \mathbb{N}^{+}$ physical quantities of index $k$ at time $t$\\
$\boldsymbol{U}_k^0$ & initial condition of index $k$ at time $t=0$\\
$\boldsymbol{u}_{k, i}^{t}\in \mathcal{R}^{N}$ & a physical quantity, such as vorticity\\
$\left\{\boldsymbol{x}_j \in \Omega\right\}_{1 \leq j \leq N}$ & the grid (also as discretized spatial coordinates) \\
$\mathcal{A} = \{\boldsymbol{a}_k\}$ &the input conditions space \\
$\boldsymbol{a}_k$ & a set input conditions of index $k$, containing initial condition, and other conditions\\
$\mathcal{S} = \{\boldsymbol{S}_k\}$ & the solutions space \\
$\boldsymbol{S}_k = (\boldsymbol{U}_k^{1}, \cdots, \boldsymbol{U}_k^{T})$ & a set output of index of index $k$\\ 
$\mathcal{D} = \{(\boldsymbol{a}_k, \boldsymbol{S}_k)\}_{1\leq k \leq D}$&
the dataset, where $\boldsymbol{S}_k = \mathcal{G}(\boldsymbol{a}_k)$\\
$\mathcal{G}:\mathcal{A}\rightarrow \mathcal{S}$ & the mapping of our goal to learn\\
$\mathcal{G}_{\theta}:\boldsymbol{a}_k\rightarrow\tilde{\boldsymbol{S}}_k, 1\leq k \leq D$ & a parameterized neural network with parameters $\theta\in \boldsymbol{\Theta}$\\
$\boldsymbol{\Theta}$ & the parameter space\\
$T$ & the time-step size for inference \\
$T'$ & the time-step size of the training dataset, where $1\leq T'\ll T$\\
\midrule
Methodology&\\
\midrule
$\tilde{\boldsymbol{U}}^t$ & the updated physical quantity by using the given boundary conditions\\
DF & Diffusive Flows \\
CF & Convective Flows \\
IST & Internal Source Term\\
EST & External Source Term\\
TSSM & Temporal-Spatial Stepping Module \\
$k_x$, $k_y$ & the spectral space dimensions\\
\midrule
Experiments&\\ 
\midrule
C Int. & the task of coefficient interpolation\\
C Ext. & the task of coefficient extrapolation\\
$\epsilon$ & the mean $L_2$ relative error \\
BC $\epsilon$ & the mean $L_2$ relative error on the boundary\\
$N_{P}$ & the number of trainable parameters\\
\bottomrule
\end{tabular}
}
\label{tab_notations}
\end{table*}

%% file: alg1.tex
\begin{algorithm}[H]
\SetAlgoLined
\textbf{Initialization:} Fixed or pre-defined convolutional kernels $K$ (with parameters $\theta$), as shown in Fig.~\ref{TSSM} (Left); Initialize other network parameters $\theta \in \Theta$\;\\
\textbf{Input:} A set of inputs $\boldsymbol{a}_k$ for $1 \leq k \leq D_{0}$, time interval $\Delta t$, and temporal trajectory length $T'$\;\\
\textbf{Output:} The mapping $\mathcal{G}_{\theta}$, where $\tilde{\boldsymbol{U}}_k$ $\leftarrow$ $\tilde{\mathcal{G}}_{\theta}(\boldsymbol{a}_k)$\;\\
\For{$k=1$ \textit{to} $D_0$}
{   
    $\boldsymbol{a}_k \leftarrow \boldsymbol{U}_k^{t=0}, \boldsymbol{\lambda}, X_F$, BCs \;\\
    \For{$t=1$ to $T'$}
    {
        $\tilde{\boldsymbol{U}}_k^{t} \leftarrow  \boldsymbol{U}_k^{t}$  \# Embedding BCs\;\\
        $(\cdot)^n$, $\nabla^n(\cdot), n=0,1,2 \leftarrow \tilde{\boldsymbol{U}}_k^{t} \circledast K$\;\\
        DF, CF $\leftarrow (\cdot)^n$, $\nabla^n(\cdot), \boldsymbol{a}_k$, $n=0,1,2$\;\\
        IST, EST $\leftarrow$ ResNet$(\tilde{\boldsymbol{U}}_k^{t}, \boldsymbol{a}_k)$\;\\
        $\frac{\partial \tilde{\boldsymbol{U}}_k^{t}}{\partial t} \leftarrow$ DF$+$CF$+$IST$+$EST\;\\
        Next states $\boldsymbol{U}_k^{t+1} \leftarrow \text{Neural ODE}(\tilde{\boldsymbol{U}}_k^{t}, \frac{\partial \tilde{\boldsymbol{U}}_k^{t}}{\partial t};\Delta t)$\;\\
        Update input $\boldsymbol{a}_k \leftarrow \boldsymbol{U}_k^{t+1}$\;
    }
    Subsequent trajectory $\tilde{\boldsymbol{U}}_k$ $\leftarrow$ $\tilde{\mathcal{G}}_{\theta}(\boldsymbol{a}_k)$ \;\\
    Loss $\mathcal{L}_r(\boldsymbol{\theta})$ $\leftarrow$ Eq.~\ref{all_lossfunction} \;\\
    Update weights $\theta$ by minimizing the loss $\mathcal{L}_r(\boldsymbol{\theta})$ \;
}
\caption{Structure-preserved localized operator.}
\label{alg1_localized}
\end{algorithm}

%% file: alg2.tex
\begin{algorithm}[H]
\SetAlgoLined
\textbf{Initialization:} E-Conv ($1 \times 1$ conv) and S-conv (spectral convolutions) with parameters $\theta$ as shown in Fig.~\ref{TSSM} (Right) \;\\
\textbf{Input:} A set of inputs $\boldsymbol{a}_k$ for $1 \leq k \leq D_{0}$, time interval $\Delta t$, and temporal trajectory length $T'$\;\\
\textbf{Output:} The mapping $\mathcal{G}_{\theta}$ where $\tilde{\boldsymbol{U}}_k$ $\leftarrow$ $\tilde{\mathcal{G}}_{\theta}(\boldsymbol{a}_k)$ \;\\
\For{$k=1$ \KwTo $D_0$}{
    $\boldsymbol{a}_k \leftarrow \boldsymbol{U}_k^{t=0}, \boldsymbol{\lambda}, X_F$, BCs\;\\
    \For{$t=1$ \KwTo $T'$}{
        $\tilde{\boldsymbol{U}}_k^{t} \leftarrow  \boldsymbol{U}_k^{t}$  \# Embedding BCs\;\\
        $k_x, k_y, \hat{\cdot} \leftarrow \text{FFT}(\tilde{\boldsymbol{U}}_k^{t})$\;\\
        $\hat{(\cdot)}^n$, $\hat{\nabla}^n(\cdot), n=0,1,2 \leftarrow \text{E-Conv}(k_x, k_y, \hat{\cdot})$\;\\
        $\text{DF}, \text{CF} \leftarrow \text{IFFT}((\cdot)^n$, $\nabla^n(\cdot), \boldsymbol{a}_k)$\;\\
        $\text{IST}, \text{EST} \leftarrow \text{S-conv}(\tilde{\boldsymbol{U}}_k^{t}, \boldsymbol{a}_k)$\;\\
        $\frac{\partial \tilde{\boldsymbol{U}}_k^{t}}{\partial t} \leftarrow \text{DF} + \text{CF} + \text{IST} + \text{EST}$\;\\
        Next states $\boldsymbol{U}_k^{t+1} \leftarrow \text{Neural ODE}(\tilde{\boldsymbol{U}}_k^{t}, \frac{\partial \tilde{\boldsymbol{U}}_k^{t}}{\partial t};\Delta t)$\;\\
        Update input $\boldsymbol{a}_k \leftarrow \boldsymbol{U}_k^{t+1}$\;
    }
    Subsequent trajectory $\tilde{\boldsymbol{U}}_k$ $\leftarrow$ $\tilde{\mathcal{G}}_{\theta}(\boldsymbol{a}_k)$ \;\\
    Compute loss $\mathcal{L}_r(\boldsymbol{\theta})$ according to Eq.~\ref{all_lossfunction} \;\\
    Update weights $\theta$ by minimizing the loss $\mathcal{L}_r(\boldsymbol{\theta})$ \;
}
\caption{Structure-Preserved Spectral Operator}
\label{alg2_spectral}
\end{algorithm}